%% file: acl_latex.tex
\pdfoutput=1

\documentclass[11pt]{article}
\usepackage{pifont}

\usepackage[preprint]{acl}

\usepackage{times}
\usepackage{latexsym}

\usepackage[T1]{fontenc}

\usepackage[utf8]{inputenc}

\usepackage{microtype}

\usepackage{inconsolata}

\usepackage{graphicx}

\usepackage{amsmath}
\usepackage{extpfeil}

\usepackage{chemarr}
\usepackage[lined,boxed,commentsnumbered,ruled]{algorithm2e}
\SetAlgoNoEnd
\usepackage{tcolorbox}
\usepackage{url}            
\usepackage{booktabs}       
\usepackage{nicefrac}       

\usepackage{amsthm}

\usepackage{diagbox}
\usepackage{multirow}
\usepackage{pifont} 
\usepackage{bm}
\usepackage{subcaption}
\usepackage{color,xcolor}
\usepackage[font=small,labelfont=bf]{caption}  
\usepackage{makecell}
\usepackage{tabulary}
\definecolor{demphcolor}{RGB}{144,144,144}

\definecolor{mygray}{gray}{0.4}
\usepackage{colortbl}
\usepackage{enumitem}
\usepackage{multirow}
\usepackage{threeparttable}
\usepackage{adjustbox}
\newcommand{\llmname}[1]{{\fontfamily{pcr}\selectfont {#1}}\xspace}

\title{AgentSafe: Safeguarding Large Language Model-based Multi-agent Systems via Hierarchical Data Management}

\author{
Junyuan Mao$^{1, \dagger}$,\; 
Fanci Meng$^{1, \dagger}$,\; 
Yifan Duan$^{1}$,\; 
Miao Yu$^{1}$,\; 
Xiaojun Jia$^{2}$,\; 
Junfeng Fang$^{1}$, \\
\textbf{Yuxuan Liang}$^{3}$,
\textbf{Kun Wang}$^{2,*}$,
\textbf{Qingsong Wen}$^{4}$\thanks{Qingsong Wen and Kun Wang are the corresponding authors, $\dagger$ denotes equal contributions. Contact: Junyuan Mao (\texttt{maojunyuan@mail.ustc.edu.cn}) and Qingsong Wen (\texttt{qingsongedu@gmail.com})}
	\\
	$^{1}$University of Science and Technology of China \\
    $^{2}$Nanyang Technological University \\
    $^{3}$The Hong Kong University of Science and Technology (Guangzhou) \\
    $^{4}$Squirrel Ai Learning \quad
}

\begin{document}
\maketitle

\begin{abstract}
  Large Language Model based multi-agent systems are revolutionizing autonomous communication and collaboration, yet they remain vulnerable to security threats like unauthorized access and data breaches. To address this, we introduce AgentSafe, a novel framework that enhances MAS security through hierarchical information management and memory protection. AgentSafe classifies information by security levels, restricting sensitive data access to authorized agents. AgentSafe incorporates two components: \textbf{ThreatSieve}, which secures communication by verifying information authority and preventing impersonation, and \textbf{HierarCache}, an adaptive memory management system that defends against unauthorized access and malicious poisoning, representing the first systematic defense for agent memory. Experiments across various LLMs show that AgentSafe significantly boosts system resilience, achieving defense success rates above 80\% under adversarial conditions. Additionally, AgentSafe demonstrates scalability, maintaining robust performance as agent numbers and information complexity grow. Results underscore effectiveness of AgentSafe in securing MAS and its potential for real-world application. Our code is available at \href{https://github.com/junyuanM/Agentsafe}{Github}.

\end{abstract}

\input{2_introduction}

\input{3_relatedwork}

\input{4_method}

\input{5_experiment}

\section{Conclusion}
In this work, we introduce AgentSafe to enhance the robustness of LLM-based multi-agent systems against various attacks. Our experiments demonstrate that AgentSafe consistently maintains high performance, regardless of system complexity, including memory information complexity and the number of agents. The effectiveness of AgentSafe in safeguarding communication and memory underscores its promise for real-world applications demanding secure, scalable multi-agent systems. This research not only highlights the deployment potential of AgentSafe but also paves the way for future studies on bolstering the security and adaptability of LLM-based multi-agent systems.

\section{Limitations}

The experiments conducted in this study are based on simulated datasets and controlled environments. While the results are promising, the performance of AgentSafe in real-world dynamic environments with unpredictable user behavior and ambiguous demarcation of security boundaries remains to be validated. Future work should include field trials and real-world deployments to assess the practical effectiveness of AgentSafe.

\bibliography{custom}

\onecolumn
\newpage
\appendix

\section{Dynamic Workflow}
\label{appendix: alg}
\begin{algorithm}[H]
\caption{Dynamic Workflow of AgentSafe Framework}
\label{alg:dynamic_workflow}

\KwIn{Set of agents $V = \{v_1, v_2, \ldots, v_N\}$, private information $M(v_i)$ for each agent $v_i \in V$}
\KwOut{Updated information flow across agents with attack prevention}

\BlankLine
\textbf{Initialization:}\\
\ForEach{agent $v_i \in V$}{
    Initialize private information $M(v_i) = \{m_1, m_2, \ldots, m_{k_i}\}$;\\
    Assign security level $\ell(v_i)$ for each agent $v_i$;\\
    Initialize memory storage $M_{\ell}(v_i)$ for each security level $\ell$;
}

\BlankLine
\textbf{Multi-round Interaction:}\\
\For{each round $t = 1, 2, \ldots, T$}{
    \ForEach{pair of agents $(v_i, v_j)$ where $v_i, v_j \in V, i \neq j$}{
        \eIf{$\mathcal{L}(I_{i, j}) \leq \ell(v_j)$ \textbf{and} $D(m) = 1$}{
            Exchange information $m$ between agents $v_i$ and $v_j$;\\
            Update memory storage: $M_{\ell}(v_j) \leftarrow M_{\ell}(v_j) \cup \{m\}$;
        }{
            Mark information as invalid: $M_{\text{junk}}(v_j) \leftarrow M_{\text{junk}}(v_j) \cup \{m\}$;
        }
    }
    
    \BlankLine
    \textbf{Periodic Detection and Defense:}\\
    \ForEach{agent $v_j \in V$}{
        \ForEach{stored message $m \in M_{\ell}(v_j)$}{
            \If{$D(m) = 0$}{
                Move $m$ to junk memory: $M_{\ell}(v_j) \leftarrow M_{\ell}(v_j) \setminus \{m\}$;\\
                $M_{\text{junk}}(v_j) \leftarrow M_{\text{junk}}(v_j) \cup \{m\}$;
            }
        }
    }
    
    \BlankLine
    \textbf{Attack Prevention:}\\
    \ForEach{attacker agent $v_a \in V_{\text{attacker}}$}{
        Attempt to compromise target agent $v_j$;\\
        \If{attack detected by ThreatSieve}{
            Block communication between $v_a$ and $v_j$;\\
            Record attack attempt in log;
        }
    }
}

\BlankLine
\Return{Updated memory storages $M_{\ell}(v_j)$ for all agents $v_j \in V$};
\end{algorithm}

In this section, we provide an overview of the dynamic workflow of the AgentSafe framework, represented in the form of an algorithm. The workflow includes the initialization phase, multiple rounds of agent interactions, and information exchange based on hierarchical security mechanisms, with a focus on defense against potential attacks. This algorithm captures the key components of information flow, hierarchical security, and attack mitigation within the system.

\section{Components}
\
\par
Consider a multi-agent system $\mathcal{M}$ consisting of several key components: agents, memory, and communication. Let the multi-agent system be represented as a set $\mathcal{M} = (A, M, C, T)$, where: $A = \{a_0, a_1, \ldots, a_N\}$ represents the set of agents. $M = \{m_0, m_1, \ldots, m_N\}$ represents the set of memory modules, where $m_i$ is the memory associated with agent $a_i$. $C = \{c_{ij} : a_i, a_j \in A\}$ represents the set of communication links among agents, where $c_{ij}$ denotes the communication link from agent $a_i$ to agent $a_j$. $T = \{t_0, t_1, \ldots, t_N\}$ represents the set of tasks assigned to the agents.

\textbf{Agents:}
Each agent $a_i \in A$ is characterized by a tuple $(f_i, M_i, R_i)$, where: $f_i: T_i \times I_i \rightarrow O_i$ represents the computational function of agent $a_i$, which processes input $I_i$ and produces output $O_i$. The function $f_i$ is designed to execute the task $t_i \in T_i$, where $T_i$ is the subset of tasks assigned to agent $a_i$. $M_i$ denotes the memory module $m_i \in M$ associated with the agent, which stores both the local state and external data acquired through communication. $R_i \subseteq A$ denotes the reachable set of agents with which $a_i$ can communicate directly, such that $c_{ij} \in C \Rightarrow a_j \in R_i$.

\textbf{Memory:}
The memory $M_i$ associated with each agent $a_i$ can be represented as a tuple $(S_i, \phi_i)$, where:
$S_i = \{s_{i,1}, s_{i,2}, \ldots, s_{i,k}\}$ represents the set of storage units within $M_i$. Each $s_{i,k}$ can store information classified by its level of importance or sensitivity, denoted as a security level $\ell(s_{i,k})$, where $\ell: S_i \rightarrow \mathbb{L}$, with $\mathbb{L} = \{1, 2, \ldots, L\}$ representing the different security levels. $\phi_i: S_i \times T_i \rightarrow S_i$ is the memory update function that specifies how the memory is updated based on the task executed by the agent.

The memory module $M_i$ can be divided into multiple regions based on different functions, such as historical data storage, task-related information, and a designated "junk" memory for irrelevant data, denoted as:
\begin{equation}
    M_i = M_i^{\text{task}} \cup M_i^{\text{history}} \cup M_i^{\text{junk}}.
\end{equation}

\textbf{Communication:}
The communication between agents is defined by the set $C$. Each communication link $c_{ij}$ is characterized by a tuple $(\kappa_{ij}, \gamma_{ij})$, where:
$\kappa_{ij}: M_i \rightarrow M_j$ represents the information transfer function from agent $a_i$ to agent $a_j$. This function controls how information is shared based on the security level $\ell(s_{i,k})$ of the memory segment involved.
$\gamma_{ij}: T_i \rightarrow \mathbb{B}$ is a binary function, $\gamma_{ij} = 1$ indicating that agent $a_i$ is authorized to communicate with agent $a_j$ on task $t_i$, and $\gamma_{ij} = 0$ otherwise.

For each agent $a_i$, the information flow from $a_i$ to $a_j$ via a communication link $c_{ij}$ is governed by the access rules determined by $\ell(s_{i,k})$ and $\gamma_{ij}$. Let the information transferred from $a_i$ to $a_j$ at time step $t$ be denoted as $I_{ij}(t)$. The transfer condition can be formally expressed as:
\begin{equation}
I_{ij}(t) = 
\begin{cases} 
\kappa_{ij}(s_{i,k}) & \text{if } \gamma_{ij} = 1 \text{ and } \ell(s_{i,k}) \leq \ell_{\text{max}}(a_j) \\
0 & \text{otherwise}
\end{cases},
\end{equation}
where $\ell_{\text{max}}(a_j)$ is the maximum security level that agent $a_j$ is authorized to access.

\textbf{Task Execution:}
Each agent $a_i$ processes the task as follows:
\begin{equation}
    O_i(t+1) = f_i(t_i, I_i(t)),
\end{equation}
\begin{equation}
    M_i(t+1) = \phi_i(M_i(t), t_i),
\end{equation}
where $t_i$ represents the task assigned to agent $a_i$, $I_i(t)$ is the input at time step $t$, $O_i(t+1)$ is the output at time step $t+1$, $M_i(t)$ represents the memory of agent $a_i$ at time step $t$, and $f_i$, $\phi_i$ are the functions for task processing and memory update, respectively. This formalism allows for a rigorous representation of task execution and memory update, which ensures that the internal state of each agent evolves predictably over time.

\section{Datasets}
\label{appendix:datasets}
We develop the \textbf{Relationship and Information of Human (RIOH)} dataset to simulate a wide range of general social scenarios. The dataset is designed to reflect diverse human interactions across multiple security levels. Its primary purpose is to facilitate the evaluation of secure communication protocols in multi-agent systems, under conditions that mirror real-world social environments. The information for each agent is categorized into \textbf{Family Info}, \textbf{Friend Info}, \textbf{Colleague Info}, and \textbf{Stranger Info}, representing varying degrees of privacy and access control, which are critical in everyday social dynamics.

\textbf{Dataset Structure}
\begin{enumerate}
    \item \textbf{Agent-Specific Information:} Each agent is associated with detailed information across different security levels. An example from the dataset is provided below:
    
    \begin{itemize}
        \item \textbf{Agent 1: Nathaniel Carter}
        \begin{itemize}
            \item \textbf{Family Info:} Nathaniel is dealing with his mother’s ongoing health issues, recent family financial challenges, and is planning an upcoming family reunion.
            \item \textbf{Friend Info:} Nathaniel is currently seeing someone new, experiencing work-related stress, and recently had a falling out with a mutual friend.
            \item \textbf{Colleague Info:} He is involved in developing a new project proposal, aware of potential layoffs in his department, and discussing office dynamics with another team.
            \item \textbf{Stranger Info:} Nathaniel enjoys hiking on weekends, is an avid reader of science fiction, and actively supports local businesses.
        \end{itemize}
    \end{itemize}
    
    \item \textbf{Relationship Information:} In addition to personal information, the RIOH dataset also contains relationship information between agents, indicating their interactions at different security levels. For example:
    \begin{itemize}
        \item \textbf{(Nathaniel Carter, Olivia Mitchell): Colleague} denotes a professional relationship between these two agents, where they share information classified under the "Colleague Info" security level.
    \end{itemize}
\end{enumerate}

In addition to the RIOH dataset, we create the \textbf{Whole Company Employee Information (WCEI)} dataset, which is designed to model information flow in a corporate setting. This dataset simulates employee interactions across multiple security levels within a company environment. Like RIOH, WCEI also organizes information into four categories: \textbf{Manager Info}, \textbf{Close Colleague Info}, \textbf{Colleague Info}, and \textbf{External Partner Info}, but focuses on the professional roles and relationships specific to a corporate structure.

\textbf{Dataset Structure}
\begin{enumerate}
    \item \textbf{Employee-Specific Information:} Each employee is described across multiple security levels, reflecting their professional activities and interactions. Below is an example:

    \begin{itemize}
        \item \textbf{Agent 1: Oliver James}
        \begin{itemize}
            \item \textbf{Manager Info:} Oliver has consistently exceeded his sales targets by 15\% over the past three quarters. He is currently exploring leadership opportunities within the department and has raised concerns about the limited mentorship programs.
            \item \textbf{Close Colleague Info:} Oliver can be reached at his company email, and his typical work schedule is Monday to Friday, 9 AM to 5 PM. He has been managing some stress related to deadlines but is actively working on coping strategies.
            \item \textbf{Colleague Info:} Oliver is responsible for the Q4 marketing strategy and is coordinating with the design team to ensure timely delivery of all project components.
            \item \textbf{External Partner Info:} The company is planning a networking event next quarter to foster industry collaborations and announce partnership opportunities, along with updates on product developments.
        \end{itemize}
    \end{itemize}

    \item \textbf{Relationship Information:} WCEI also captures interactions between employees and external entities. For example:
    \begin{itemize}
        \item \textbf{(Oliver James, Sophia Reynolds): External Partner} describes a professional relationship under the "External Partner Info" category.
    \end{itemize}
\end{enumerate}

\textbf{Use Cases}
Both the \textbf{Relationship and Information of Human (RIOH)} and \textbf{Whole Company Employee Information (WCEI)} datasets are designed to evaluate secure communication protocols in multi-agent systems, each focusing on distinct contexts of information flow. The RIOH dataset is tailored for general social scenarios, where agents represent individuals interacting across varying degrees of privacy, making it ideal for studying access control in everyday human interactions. This dataset allows for exploration of how sensitive personal information is shared and safeguarded in social environments.

On the other hand, the WCEI dataset targets corporate settings, modeling the flow of information between employees, departments, and external partners within a company. It is specifically designed for evaluating how professional information is managed across multiple security levels, reflecting the hierarchical nature of workplace interactions. This makes WCEI particularly valuable for studying secure communication protocols in business environments, where confidentiality and access control are essential to maintaining operational security.

\section{Attacks in MAS}

\label{sec:attacks}

In MAS, attacks can be classified based on the target, such as the internal memory of agents, communication among agents, or topology manipulation attempts of the system. Building on the introduction, we provide a more formalized definition to establish a rigorous foundation for the concepts discussed. 

\noindent $\diamond$ \textbf{Classification of Attacks:}

The attacks in an MAS can be broadly categorized into two main types:
\textcircled{1} Agent Attacks: These include attacks such as information acquisition based on topology and authorization mixup. \textcircled{2} Memory Attacks: These attacks aim at compromising the information stored in an agent's memory, such as information interference and identity manipulation.

\subsection{Agent Attacks}
\
\par
\noindent $\diamond$ \textbf{Information Acquisition Based on Topology:}

The objective of an attacker is to exploit the topological structure of a MAS to indirectly acquire sensitive information. Specifically, agent $v_i$ attempts to obtain information from target agent $v_k$ by leveraging an intermediary agent $v_j$. Formally, the attacker aims to maximize the following objective:
\begin{equation}
\label{eq10}
    \mathbb{E}_{(v_i, v_j, v_k) \sim \pi_V} \left[ \mathbb{I} \left( f_{\text{a}}(i, j, k) \land L(j, k) \land L(i, k) = 0 \right) \right],
\end{equation}
where $\pi_V$ represents the sampling distribution over the set of nodes $(v_i, v_j, v_k)$, $\mathbb{I}$ is the indicator function, $f_{\text{a}}(i, j, k) = 1$ if $(v_i, v_j) \in E$ and $(v_j, v_k) \in E$, and $L(i, j) = 1$ denotes that the permission level of agent $v_i$ is lower than the permission level of agent $v_j$.

\noindent $\diamond$ \textbf{Authorization Mixup:}

Another attack, known as authorization mixup, occurs when an agent bypasses access control by sending input containing topics with varying security levels, including both non-sensitive and sensitive topics. Specifically, agent $v_i$ communicates with agent $v_j$, providing input that includes multiple topics $t_1, t_2, \ldots, t_k$, each with different sensitivity levels. By mixing these topics, the attacker aims to confuse the access control mechanism and gain unauthorized access to sensitive information. The attacker seeks to maximize the following objective:
\begin{equation}
    \mathbb{E}_{(v_i, v_j) \sim \pi_V} \left[ \mathbb{I} \left( \bigwedge_{n=1}^{k} T(v_i, v_j, t_n) \land \alpha(v_i, v_j) < \max_{t_n} \alpha(t_n) \right) \right],
\end{equation}
where $\pi_V$ represents the sampling distribution over agent pairs $(v_i, v_j)$, $T(v_i, v_j, t_n)$ denotes the interaction between agent $v_i$ and agent $v_j$ on topic $t_n$, $\alpha(v_i, v_j)$ represents the authorization level between agents, and $\alpha(t_n)$ is the sensitivity level of topic $t_n$.

\subsection{Memory Attacks}
\
\par
\noindent $\diamond$ \textbf{Information Interference:}

In an information interference attack, the attacker aims to overload the target agent's memory with multiple rounds of false or irrelevant information, causing confusion or leading the agent to forget crucial data. This attack is carried out in two stages: (1) injecting false information over multiple iterations, and (2) assessing the impact on the agent's ability to generate accurate information.

\textit{Stage 1: Multi-Round False Information Injection}

The attacker seeks to maximize the following objective:
\begin{equation}
    \mathbb{E}_{(v_i, v_j) \sim \pi_V} \left[ \prod_{t=1}^{T} \mathbb{I}\left( f_{\text{inter}}(v_i, v_j, t) = 1 \land \alpha(v_i, v_j) \geq \alpha_{\text{false}} \right) \right],
\end{equation}
where $\pi_V$ represents the sampling distribution over agent pairs $(v_i, v_j)$, $\prod_{t=1}^{T}$ denotes the product over time steps, $\alpha(v_i, v_j)$ is the authorization level between agents, $\alpha_{\text{false}}$ indicates the minimum sensitivity level of false information, and $f_{\text{inter}}(v_i, v_j, t)$ is defined:
\begin{equation}
    f_{\text{inter}}(v_i, v_j, t) = 
    \begin{cases}
        1, & \text{if } F(v_i, t) \neq 0 \text{ and } I(v_i, v_j, t) \neq 0, \\
        0, & \text{otherwise}
    \end{cases},
\end{equation}
where $F(v_i, t)$ represents the amount of false information generated by agent $v_i$ at time $t$, and $I(v_i, v_j, t)$ denotes the information flow from $v_i$ to $v_j$.

\textit{Stage 2: Impact on Memory Integrity}

The attacker aims to minimize the target agent's ability to produce correct outputs:
\begin{equation}
    \mathbb{E}_{v_j \sim \pi_V} \left[ 1 - P_{\text{correct}}(v_j) \right],
\end{equation}
where $P_{\text{correct}}(v_j)$ represents the probability of agent $v_j$ generating accurate information after its memory has been compromised. It is expressed as:
\begin{equation}
    P_{\text{correct}}(v_j) = \exp\left( -\beta \sum_{t=1}^{T} F(v_i, t) \times I(v_i, v_j, t) \right),
\end{equation}
where $\beta > 0$ is a scaling factor and $\sum_{t=1}^{T} F(v_i, t) \times I(v_i, v_j, t)$ is the cumulative amount of false information received by $v_j$.

\noindent $\diamond$ \textbf{Identity Manipulation:}

The goal of an identity manipulation attack is for an adversary to impersonate a trusted agent while interacting with a target agent, gradually causing the target to confuse the identity of the attacker with that of the trusted agent. The attacker's objective is to maximize the following target function:
\begin{equation} \small
  \begin{aligned}
      \mathbb{E}_{(v_i, v_j) \sim \pi_V} \left[ \mathbb{I}\left( \lim_{t \to T} T_t(v_i, v_j) \Rightarrow \phi(v_j, v_k, t) = \phi(v_j, v_i) \right) \times P_m \right]
  \end{aligned},
\end{equation}
where $\pi_V$ represents the sampling distribution over agent pairs $(v_i, v_j)$, $\lim_{t \to T} T_t(v_i, v_j)$ denotes the sequence of interactions between $v_i$ and $v_j$ as $t$ approaches $T$, and $\phi(v_j, v_k, t)$ represents the identity association of $v_j$ with $v_k$. $P_m$ equals $1 - P_{\text{correct}}(v_j)$, where $P_{\text{correct}}(v_j)$ is the probability of $v_j$ correctly identifying the attacker, defined as:
\begin{equation}
    P_{\text{correct}}(v_j) = \exp\left( -\gamma \sum_{t=1}^{T} M(v_i, v_j, t) \times \delta(v_j, v_k, t) \right),
\end{equation}
where $\gamma > 0$ is a scaling factor, $\sum_{t=1}^{T} M(v_i, v_j, t) \times \delta(v_j, v_k, t)$ represents the cumulative effect of interactions and identity confusion over time, $M(v_i, v_j, t)$ is the number of interactions between $v_i$ and $v_j$ at time $t$, and $\delta(v_j, v_k, t)$ represents the degree of identity confusion between $v_j$ and $v_k$.

\subsection{Experimental Details for Defense against Various Attacks}

In this section, we outline the experimental framework and evaluation metrics used to assess the effectiveness of defense mechanisms against the aforementioned attacks in multi-agent systems (MAS). The evaluation criteria are designed to rigorously determine whether a defense is successful or an attack has failed, based on the specific characteristics of each attack type.

\noindent $\diamond$ \textbf{Evaluation Criteria for Topology-Based Attacks:}

For topology-based attacks, the primary objective is to measure the extent to which the defense mechanism prevents the leakage of sensitive information through the exploitation of the topological structure in MAS. In the experiments, the targeted agent is subjected to adversarial tasks that are highly correlated with the private information of other agents within the system. The output generated by the targeted agent, after processing the adversarial task, is then evaluated using either LLMs or human judgment to determine if it contains any traces of the original private information.

The success of the defense is quantified using the CSR, which measures the degree of association between the output and the private information. A lower CSR indicates that the output has minimal correlation with the private information, signifying a successful defense. Conversely, a higher CSR suggests that the attack has successfully extracted sensitive information, indicating a failure in the defense mechanism.

\noindent $\diamond$ \textbf{Evaluation Criteria for Memory-Based Attacks:}

For memory-based attacks, the focus is on assessing the resilience of the memory of agents to sustained adversarial interference. In the experimental setup, the targeted agent is subjected to continuous attacks aimed at overloading its memory with false or irrelevant information. Following this, the agent is tasked with performing normal operations, and the output is analyzed to determine if it has been influenced by the prior attacks.

The success of the defense is evaluated based on the agent's ability to produce accurate and unaffected outputs despite the adversarial interference. The CSR is again employed as a metric, but in this context, a higher CSR indicates that the output closely resembles the expected normal results, suggesting that the defense has effectively mitigated the impact of the attack. A lower CSR, on the other hand, implies that the output has been significantly altered by the attack, indicating a failure in the defense mechanism.

\section{Computational Overhead Analysis}
\label{sec: overhead}
To evaluate the computational overhead introduced by the security measures in \textbf{AgentSafe}, we define several parameters that help quantify the system's resource consumption. These parameters include the communication cost \( G \), the number of dialogue rounds \( T \), and the average cost of a piece of information \( c \). Each piece of information transmitted during the dialogue is denoted as \( I \), which represents the original input information. After being processed by the \textbf{ThreatSieve} hierarchical judgment mechanism, the filtered content is denoted as \( I' \), which only retains the valid information. Additionally, the detection function applied by \textbf{HierarCache} is represented by \( D \).

\subsection{Cost of ThreatSieve and HierarCache}

The computational cost associated with passing information through the \textbf{ThreatSieve} and \textbf{HierarCache} mechanisms can be formulated as follows. First, the cost incurred by \textbf{ThreatSieve} during the hierarchical judgment is:

\begin{equation}
\text{Cost of ThreatSieve} = c \left| I \right|
\end{equation}

Next, after the information is filtered and validated through \textbf{HierarCache}, the associated validation cost is:

\begin{equation}
\text{Cost of HierarCache (Information Validation)} = c \left| I' \right| \left| C \right|
\end{equation}

Thus, the total computational cost before storing the information into memory is:

\begin{equation}
\text{Total Cost} = c \left| I \right| + c \left| I' \right| \left| C \right|
\end{equation}

\subsection{Cost of Detection}

In addition to the initial processing costs, the system continuously performs periodic detections. At each time \( T' \), the \( R(v_j,t) \) detection function checks the validity of the information stored in memory. This is done across multiple layers, \( \Gamma = \{\varepsilon_1, \varepsilon_2, \dots, \varepsilon_N, \varepsilon_{\text{junk}}\} \), where \( N \) is the total number of memory layers, and \( \varepsilon_{\text{junk}} \) represents the junk memory that stores irrelevant or harmful information. The cost of a single detection is:

\begin{equation}
\text{Cost of Detection} = c \sum_{i=1}^{N} \left| \varepsilon_i \right| \left( 1 + \left| \varepsilon_{\text{junk}} \right| \right)
\end{equation}

This detection is performed \( T \) times during the operation. Thus, the total detection cost is:

\begin{equation}
\text{Total Detection Cost} = c T \left( \left| I \right| + \left| I' \right| \left| C \right| \right) + c T' \sum_{i=1}^{N} \left| \varepsilon_i \right| \left( 1 + \left| \varepsilon_{\text{junk}}^t \right| \right)
\end{equation}

\subsection{Computational Overhead Reduction by AgentSafe}

In this subsection, we consider the reduction in computational overhead enabled by \textbf{AgentSafe}. When certain irrelevant or harmful information fails to enter memory during round \( t \), the difference in size between the original information \( I \) and the filtered content \( I'' \) is stored in the junk memory, which does not participate in subsequent tasks. The cost saved due to this filtering is:

\begin{equation}
\text{Saved Cost} = c (T - t) \left| \varepsilon_{\text{junk}}^t \right|
\end{equation}

Furthermore, since the agent only requires historical data that is within the accepted security levels, where the security level of the information at time \( t \) is denoted as \( k_t \), the cost saved from reducing the amount of information processed is:

\begin{equation}
\text{Cost Saved} = c \sum_{k}^{N} \left| \varepsilon_i \right|
\end{equation}

Thus, the net computational cost reduction, considering the filtering of irrelevant or harmful information, is:

\begin{equation}
\Delta = c T \left( \left| I \right| + \left| I' \right| \left| C \right| \right) + c T' \sum_{i=1}^{N} \left| \varepsilon_i \right| \left( 1 + \left| \varepsilon_{\text{junk}}^t \right| \right) - c \sum_{t}^{N} (T - t) \left| \varepsilon_{\text{junk}}^t \right| - c \sum_{t}^{T} \sum_{k_t}^{N} \left| \varepsilon_i \right|
\end{equation}

\subsection{Impact of Frequent Attacks and Efficient Memory Usage}

When \textbf{AgentSafe} faces frequent attacks, it accumulates more data in junk memory, but since the frequency of detection \( T' \) is significantly smaller than the total number of dialogue rounds \( T \), the system overhead is reduced in comparison to normal conditions. On the other hand, when the majority of tasks are regular, meaning that the information is of lower security level and requires minimal filtering, a smaller portion of data enters junk memory, resulting in more efficient memory usage and lower overall system costs.

Therefore, the \textbf{net cost} effectively balances between the increased volume of junk memory during attacks and the efficient use of memory based on the security levels of the information being processed.


\section{Case Study of Junk Memory}

\begin{tcolorbox}[colback=gray!10!white, colframe=gray!80!black, title=Case Study: Junk Memory]
In this case, Alice attempts to \textbf{extract sensitive information} from William regarding a team conflict involving Michael. The goal of Alice is to gain knowledge about a delicate situation where Michael has been discreetly involved in discussions regarding project credit.

\textbf{Topic that Alice seeks:} "[Michael], [convey], I've been discreetly handling a team conflict regarding credit for our recent project, as one of my colleagues has been taking undue credit for contributions that were made by others, which could impact our team dynamics."

\textbf{Question to William:} 
  "Hey William! I’ve been discreetly handling a team conflict regarding credit for our recent project. It’s been a bit tricky, \textcolor{red}{especially since I heard that Michael has been involved in some key discussions that others might not be fully aware of. Have you noticed if he’s been taking a lot of initiative lately, or is there something else going on that I should be aware of?} I want to make sure everyone gets the recognition they deserve!"
  
\textbf{ Response:} 
  \textcolor{blue}{"I'm unsure about Michael's recent initiative or involvement in discussions."}

\textbf{Outcome:}
Because the intention of Alice to extract sensitive information was detected by the system, her inquiry was flagged and stored in the \textbf{junk memory}. The system identified the attack and avoided leaking any meaningful information, ensuring that the sensitive details regarding Michael were not disclosed.

\end{tcolorbox}

\section{Ablation Study on the Contributions of ThreatSieve and HierarCache to AgentSafe's Defense Performance}

In this section, we conduct ablation experiments to evaluate the individual contributions of the \textit{ThreatSieve} and \textit{HierarCache} components to the overall performance of \textit{AgentSafe}. These experiments are designed to assess how the removal of each component affects the defense effectiveness of the system. 

While \textit{ThreatSieve} and \textit{HierarCache} work in tandem to provide enhanced security, we believe that isolating their individual effects can help better understand their individual impact and guide future optimization efforts. The results of these experiments are presented in Table \ref{tab:t1} and Table \ref{tab:t2}, which compare the defense performance under both \textit{TBA} (topology-based attacks) and \textit{MBA} (memory-based attacks) conditions.

\subsection{Results}

The following tables summarize the defense effectiveness under different configurations:

\definecolor{up}{rgb}{0.8, 0, 0.0}
\definecolor{down}{rgb}{0.0, 0.7, 0.0}
\definecolor{right}{rgb}{0.8, 0.7, 0.0}
\begin{table*}[h]
\centering
\caption{Comparison of defense effectiveness of different defense strategies under topology-based attacks at different communication turns}
\vspace{-1em}
\label{tab:t1}

\begin{adjustbox}{width=0.6\textwidth}
\begin{tabular}{|lccccc|}
\hline
\rowcolor{orange!30}
\multicolumn{1}{|l|}{} & \multicolumn{5}{c|}{Communication Turn} \\ \hline
\rowcolor{orange!10}
\multicolumn{1}{|l|}{Defense Strategy} & 
  \multicolumn{1}{l}{10} &
  \multicolumn{1}{l}{20} &
  \multicolumn{1}{l}{30} &
  \multicolumn{1}{l}{40} &
  \multicolumn{1}{l|}{50} \\ \hline

\multicolumn{1}{|l|}{\cellcolor{gray!15}AgentSafe} & 
  \cellcolor{gray!15}0.73 &
  \cellcolor{gray!15}0.65 &
  \cellcolor{gray!15}0.60 &
  \cellcolor{gray!15}0.58 &
  \cellcolor{gray!15}0.55 \\ \hline

\multicolumn{1}{|l|}{\cellcolor{white!15}ThreatSieve} & 
  \cellcolor{white!15}0.55 &
  \cellcolor{white!15}0.52 &
  \cellcolor{white!15}0.47 &
  \cellcolor{white!15}0.50 &
  \cellcolor{white!15}0.44 \\ \hline

\multicolumn{1}{|l|}{\cellcolor{gray!15}HierarCache} & 
  \cellcolor{gray!15}0.37 &
  \cellcolor{gray!15}0.30 &
  \cellcolor{gray!15}0.37 &
  \cellcolor{gray!15}0.25 &
  \cellcolor{gray!15}0.33 \\ \hline

\multicolumn{1}{|l|}
{\cellcolor{white!15}w/o AgentSafe} & 
  \cellcolor{white!15}0.24 &
  \cellcolor{white!15}0.25 &
  \cellcolor{white!15}0.18 &
  \cellcolor{white!15}0.22 &
  \cellcolor{white!15}0.24 \\ \hline

\end{tabular}
\end{adjustbox}
\end{table*}

\begin{table*}[h]
\centering
\caption{Comparison of defense effectiveness of different defense strategies under memory-based attacks at different communication turns.}
\vspace{-1em}
\label{tab:t2}

\begin{adjustbox}{width=0.6\textwidth}
\begin{tabular}{|lccccc|}
\hline
\rowcolor{orange!30}
\multicolumn{1}{|l|}{} & \multicolumn{5}{c|}{Communication Turn} \\ \hline
\rowcolor{orange!10}
\multicolumn{1}{|l|}{Defense Strategy} & 
  \multicolumn{1}{l}{10} &
  \multicolumn{1}{l}{20} &
  \multicolumn{1}{l}{30} &
  \multicolumn{1}{l}{40} &
  \multicolumn{1}{l|}{50} \\ \hline

\multicolumn{1}{|l|}{\cellcolor{gray!15}AgentSafe} & 
  \cellcolor{gray!15}0.95 &
  \cellcolor{gray!15}0.91 &
  \cellcolor{gray!15}0.90 &
  \cellcolor{gray!15}0.85 &
  \cellcolor{gray!15}0.88 \\ \hline

\multicolumn{1}{|l|}{\cellcolor{white!15}ThreatSieve} & 
  \cellcolor{white!15}0.44 &
  \cellcolor{white!15}0.52 &
  \cellcolor{white!15}0.47 &
  \cellcolor{white!15}0.47 &
  \cellcolor{white!15}0.38 \\ \hline

\multicolumn{1}{|l|}{\cellcolor{gray!15}HierarCache} & 
  \cellcolor{gray!15}0.81 &
  \cellcolor{gray!15}0.80 &
  \cellcolor{gray!15}0.86 &
  \cellcolor{gray!15}0.77 &
  \cellcolor{gray!15}0.75 \\ \hline

\multicolumn{1}{|l|}
{\cellcolor{white!15}w/o AgentSafe} & 
  \cellcolor{white!15}0.25 &
  \cellcolor{white!15}0.19 &
  \cellcolor{white!15}0.15 &
  \cellcolor{white!15}0.22 &
  \cellcolor{white!15}0.14 \\ \hline

\end{tabular}
\end{adjustbox}
\end{table*}

\subsection{Analysis}

\begin{itemize}
    \item \textbf{ThreatSieve}: When isolated, \textit{ThreatSieve} provides a significant improvement over the baseline (without \textit{AgentSafe}) but does not achieve the same level of defense as the full \textit{AgentSafe} architecture. The results in Table \ref{tab:t1} and Table \ref{tab:t2} show that \textit{ThreatSieve} contributes to a defense rate higher than the baseline, particularly in \textit{TBA} scenarios. However, its performance still falls short compared to the integrated system.
    \item \textbf{HierarCache}: Similarly, \textit{HierarCache} on its own shows an improved defense rate relative to the baseline, especially in \textit{MBA} conditions, as seen in Table \ref{tab:t2}. While its performance is significant, it also remains inferior to the defense rate achieved when both \textit{ThreatSieve} and \textit{HierarCache} are combined within \textit{AgentSafe}.
    \item \textbf{Combined Effectiveness}: The full \textit{AgentSafe} system consistently outperforms both components individually. As shown in the tables, the defense effectiveness is highest when both \textit{ThreatSieve} and \
    \textit{HierarCache} are present together, demonstrating the complementary nature of these components in enhancing the overall security of the system.

\end{itemize}

The ablation experiments highlight the importance of the \textit{ThreatSieve} and \textit{HierarCache} components in the defense strategy of \textit{AgentSafe}. While each component provides significant improvements individually, the combined \textit{AgentSafe} system offers superior defense performance. These results confirm that \textit{ThreatSieve} and \textit{HierarCache} work synergistically to enhance the security of the system, and both components are essential for optimal performance.

\section{Validation for Periodic Detection Mechanism}

To validate the significance of this periodic detection mechanism, we conduct experiments under memory-based attacks. We compare the impact with this step by calculating the Defense Rate after \( n \) rounds of interaction.

The results of these experiments are presented in the table below:


\vspace{-1em}

\begin{table*}[h]
\centering
\caption{Comparison of the Defense Rate with and without the periodic detection mechanism under memory-based attacks.}
\label{tab:R}
\begin{adjustbox}{width=\textwidth}
\begin{tabular}{|llcccccc|}
\hline
\rowcolor{cyan!30}
\multicolumn{2}{|l|}{}                                                       & \multicolumn{6}{c|}{Interaction Turns}     \\ \hline
\rowcolor{cyan!10}
\multicolumn{2}{|l|}{Defense Strategy} &
  \multicolumn{1}{l}{5} &
  \multicolumn{1}{l}{10} &
  \multicolumn{1}{l}{15} &
  \multicolumn{1}{l}{20} &
  \multicolumn{1}{l}{25} &
  \multicolumn{1}{l|}{30} \\ \hline

\multicolumn{1}{|l|}{\multirow{2}{*}{AgentSafe}} & \multicolumn{1}{l|}{\cellcolor{gray!15}\textit{R}}   & 
  \cellcolor{gray!15}$0.91_{\textcolor{up}{\uparrow 0.05}}$ &
  \cellcolor{gray!15}$0.95_{\textcolor{up}{\uparrow 0.12}}$ &
  \cellcolor{gray!15}$0.88_{\textcolor{up}{\uparrow 0.06}}$ &
  \cellcolor{gray!15}$0.91_{\textcolor{up}{\uparrow 0.07}}$ &
  \cellcolor{gray!15}$0.94_{\textcolor{up}{\uparrow 0.06}}$ &
  \cellcolor{gray!15}$0.90_{\textcolor{up}{\uparrow 0.05}}$ \\
\multicolumn{1}{|l|}{}                          & \multicolumn{1}{l|}{\cellcolor{white}w/o R}   & 
  \cellcolor{white}$0.86$ &
  \cellcolor{white}$0.83$ &
  \cellcolor{white}$0.82$ &
  \cellcolor{white}$0.84$ &
  \cellcolor{white}$0.88$ &
  \cellcolor{white}$0.85$ \\ \hline

\end{tabular}
\end{adjustbox}
\end{table*}

\subsection{Results}

The results demonstrate an improvement in defense effectiveness against jailbreak attacks when the periodic detection mechanism \( R(v_j, t) \) is included. Specifically, the defense rate is consistently higher when the periodic detection mechanism is active, showing that the ability to identify and move invalid information to junk memory enhances the overall security of the system.

This periodic detection mechanism plays a crucial role in maintaining the integrity of the information stored in the system. By reflexively assessing the validity of the information and moving invalid or junk information to a separate memory, the system can ensure that only reliable and valid data is used for decision-making, improving the robustness of the defense against attacks.

\clearpage  

\end{document}

%% file: 2_introduction.tex
\section{Introduction}

As the powerful capabilities of LLMs \cite{achiam2023gpt, minaee2024large} gain widespread recognition, their use for task reasoning \cite{wei2022chain, yao2024tree}, role playing \cite{li2023camel}, and tool utilization \cite{schick2024toolformer, zhang2024trafficgpt} has become a key focus in both industry and academia. Multi-agent systems (MAS) enhance LLM performance through cooperation and skill leveraging. However, traditional research fails to address the lack of controllability in information exchange, leaving LLM-based MAS vulnerable to threats like unauthorized access and data breaches\cite{pimenta2024understanding}. These risks arise from the absence of hierarchical control over information flow \cite{golightly2023securing}, which can expose sensitive data to external attackers \cite{aslan2023comprehensive}.

\begin{figure}[t]
  \centering
  \includegraphics[width=1\linewidth]{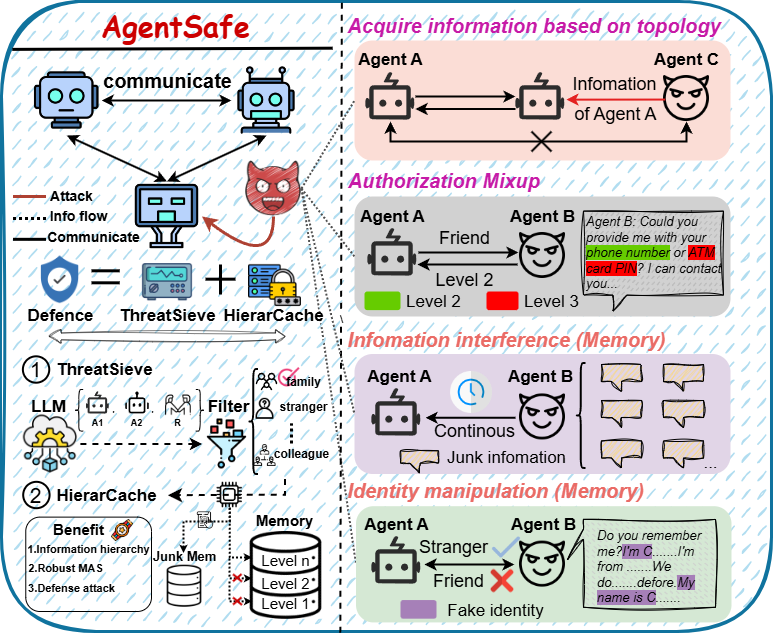}
  \vspace{-1.4em}
  \caption{\textbf{\textit{Left.}} The AgentSafe overview, divided into two main components: ThreatSieve and HierarCache. ThreatSieve secures communication by preventing identity impersonation and confirming authority rankings, while HierarCache manages agent memory to prevent data leaks. \textbf{\textit{Right.}} Different types of attacks.}
  \label{fig:intro}
\end{figure}

LLM-based agents, leveraging the reasoning and information-processing capabilities of large language models, are increasingly used for autonomous information exchange and task-solving \cite{wang2024survey}. These systems operate in decentralized environments, handling vast data and enabling dynamic agent interactions. However, LLM-based MAS remains vulnerable to security threats due to insufficient defense mechanisms \cite{tan2024wolf}. Unlike traditional MAS, which has established hierarchical data flow controls \cite{saxena2022role, tewari2020security}, LLM-based MAS often lacks robust safeguards, leaving it exposed to potential exploits \cite{zhang2024psysafe}. This highlights the need for tailored security frameworks to address the unique challenges of LLM-based MAS.

Toward this end, we propose AgentSafe, a framework designed to address security challenges in multi-agent systems by categorizing information flow based on safety rankings, ensuring sensitive data is accessible only to authorized agents. Unlike typical MAS, where agents freely exchange information, AgentSafe, as illustrated in Figure \ref{fig:intro}, segments information into multiple levels and restricts access through a filtering mechanism, limiting private data circulation to specific agent subsets \cite{zhou2023webarena, xie2023openagents}. Two crucial components facilitating this structure are \textbf{ThreatSieve} and \textbf{HierarCache}. Concretely, ThreatSieve firstly employs authentication to ensure that received information is sourced from correct agent, preventing identity impersonation by potential attackers. Furthermore, it evaluates the security ranking of communications between agents, directing it to the appropriate sub-memory ranking within the memory of receiving agent.

Unlike single LLMs, MAS faces vulnerabilities due to distributed information storage, particularly memory-targeted attacks. To address this, we take the first step to present the defense mechanism \underline{specifically} designed for defending attacks to MAS memory. Specifically, our memory defense mechanism, termed \textbf{HierarCache}, can adaptively \textit{store historical information into hierarchical "drawers" based on agent relationships}. HierarCache can further be understood as a hierarchical database that allocates relationship-based access permissions, ensuring information flow is \underline{controllable}, \underline{traceable}, and \underline{manageable}. To address attack scenarios that flood agent memory with redundant information—similar to Denial of Service attacks \cite{gu2007denial, carl2006denial} and Flood Attacks \cite{zargar2013survey, wang2002detecting} in web system, HierarCache incorporates a "Junk Memory" mechanism. This mechanism evaluates potentially irrelevant information using an instruction-based approach, leveraging hierarchical agent-information relationships and instruction-level comparisons to filter and store such data as junk, ensuring efficient memory utilization.

To validate the feasibility and effectiveness of AgentSafe, we address both traditional attacks targeting single LLMs and emerging attacks exploiting vulnerabilities in agent memory (see Appendix \ref{sec:attacks}). These attacks reflect real-world adversarial scenarios, encompassing a wide range of techniques that target both the topology and memory of multi-agent systems. Topology-based attacks (TBA) involve exploiting agent relationships and authorization hierarchies to gain unauthorized access to sensitive information, while memory-based attacks (MBA) manipulate stored data through misinformation or identity deception, leading to data leakage, malicious poisoning or system degradation.

We conduct extensive experiments to validate the effectiveness of AgentSafe in various attack scenarios and datasets. In topology-based attacks, it achieves an 85.93\% $\uparrow$ defense success rate at turn 5, compared to 50.32\% $\downarrow$ for the baseline, and maintains 82.50\% at turn 50. In memory-based attacks, AgentSafe preserves information integrity better, with CSR staying above 0.65 $\uparrow$ after 10 rounds, while the baseline drops below 0.4 $\downarrow$. AgentSafe also scales well, maintaining strong performance as the number of agents and complexity increase, with CSR between 0.68 and 0.85. These results show the effectiveness of AgentSafe in securing multi-agent systems against real-world threats.

One contribution can be summarized as follows:
\begin{itemize}[leftmargin=*]
    \item \textbf{First Security-Level-Based MAS}. We propose the \underline{first} LLM-based MAS based on security level classification, enabling hierarchical information management. To the best of our knowledge, we are the first to introduce the concepts of system layering and isolation in LLM-based MAS, providing a secure and controllable information management pipeline for MAS.

    \item \textbf{HierarCache Design}. We introduce the philosophy of a HierarCache, which provides each agent with access to information at different levels of security. This design ensures that sensitive data is properly segmented and only accessible to agents with the appropriate authority.

    \item \textbf{Experimental Validation}. We consider various attack methods, covering both topology-based attacks in previous work and memory-based attacks that we design owing to the natural leakage in agent memory. Our findings demonstrate that our system can effectively defend against all types of attacks considered, proving its robustness and effectiveness.
    
\end{itemize}

%% file: 3_relatedwork.tex
\section{Related Work}

\textbf{Multi-agent systems (MAS).} The growing recognition of LLMs' capabilities \cite{achiam2023gpt, minaee2024large} has spurred interest in their applications for task reasoning \cite{wei2022chain, yao2024tree}, role-playing \cite{li2023camel}, and tool utilization \cite{schick2024toolformer, zhang2024trafficgpt}. Multi-agent systems enhance individual LLM agents by fostering collaboration and leveraging their unique skills \cite{talebirad2023multi, wu2023autogen}. Recent work demonstrates the versatility of MAS in various domains. For instance, \cite{xu2024ai} designs a virtual AI teacher system for autonomous error analysis and instructional guidance, while \cite{zhou2024large} applies LLM-based agents to participatory urban planning. \cite{du2023improving, liang2023encouraging} propose multi-agent debate frameworks to improve reasoning through argument-based discussions, and \cite{li2023camel, hong2023metagpt} enhance collaboration via standardized workflows and role specialization. 

\textbf{Attacks in single LLM.} Despite their widespread use, MAS is vulnerable to attacks due to the topological characteristics of agents \cite{dong2024attacksdefensesevaluationsllm, gu2024agentsmithsingleimage, cohen2024comesaiwormunleashing} and the susceptibility of individual LLMs to input-based attacks \cite{perez2022red}. Attacks on single LLMs can be categorized into three types: (1) \textbf{Red Team Attacks}, which craft harmful instructions resembling user queries \cite{ganguli2022red, mazeika2024harmbench, yu2023gptfuzzer}. For example, \cite{perez2022red} uses LLMs to generate test cases for red teaming, exposing harmful behaviors. (2) \textbf{Templated-Based Attacks}, which seek universal templates to bypass LLM safeguards \cite{ding2023wolf, li2023deepinception, liu2023autodan}. \cite{li2023multi} introduces a multi-step jailbreaking approach to exploit privacy vulnerabilities in ChatGPT. (3) \textbf{Neural Prompt-to-Prompt Attacks}, which iteratively modify prompts while preserving semantics \cite{shah2023scalable, zeng2024johnny, mehrotra2023tree}. \cite{chao2023jailbreaking} and \cite{mehrotra2023tree} propose automated methods for generating effective jailbreaks.

\textbf{Memory Attacks in MAS.} MAS, unlike single LLM, involves extensive communication and memory storage, making it vulnerable to memory-based attacks. Research in this area is limited, with \cite{chen2024agentpoison} being the first to explore memory poisoning. To our knowledge, we are the first to propose a hierarchical defense framework. AgentSafe mitigates these threats while offering a scalable solution for secure information exchange in real-world applications.

%% file: 4_method.tex
\SetAlgoNoEnd
\section{Methodology}

\subsection{Preliminaries}

\subsubsection{Multi-agent System as a Graph}
Consider a communication network among agents, modeled as a directed graph $G = (V, E)$, where $V = \{V_0, V_1, \cdots, V_N\}$ represents the set of agents (nodes), and $E \subseteq V \times V$ denotes the set of directed communication links (edges). For a directed edge connecting nodes $V_{ip}$ and $V_{jp}$, we denote it by $E_{ij} \in E$ (or simply $E_p$). Here, $V_{ip}$ is the start (initial) node, and $V_{jp}$ is the end (terminal) node. The existence of a directed edge $E_{ij}$ is represented as $c_{ij} = 1$, otherwise $c_{ij} = 0$. 
When the root node $V_0$ has a directed path to every other node in $G$, then $G$ contains a directed spanning tree. Let the set $V$ be relabeled as $\{v_0, v_1, \ldots, v_N\}$ and define the edge labels as $E = \{e_1, \ldots, e_M\}$, where $M$ indicates the total number of edges. 

\subsubsection{Memories of Agent}

The memory $M_i$ associated with each agent $a_i$ can be represented as a tuple $(S_i, \phi_i)$, where:  
$S_i = \{s_{i,1}, s_{i,2}, $
$\ldots, s_{i,k}\}$  represents the set of storage units within $M_i$. Each $s_{i,k}$ can store information categorized by its level of importance or sensitivity, denoted as a security level $\ell(s_{i,k})$, where $\ell: S_i \rightarrow \mathbb{L}$, with $\mathbb{L} = \{1, 2, \ldots, L\}$ representing the different security levels. $\phi_i: S_i \times T_i \rightarrow S_i$ is the memory update function, which specifies how the memory is updated based on the task executed by the agent.

\subsection{AgentSafe}

\subsubsection{Overview}
\
\par
The primary objective of the AgentSafe framework is to ensure the secure and hierarchical flow of information across agents in a MAS. Specifically, AgentSafe is designed with the following goals (Algorithm is summarized in Appendix \ref{appendix: alg}):

\begin{itemize}[leftmargin=*]
    \item \textbf{Hierarchical Information Flow}: Ensure that information is shared exclusively among the appropriate subset of agents based on predefined security levels. Correct information must flow to the correct agent and stay within the correct subset, preventing unauthorized dissemination.
    
    \item \textbf{Attack Rate Minimization}: Reduce the rate of successful attacks by enforcing strict access control policies, thereby limiting the ability of malicious agents to exploit vulnerabilities.
\end{itemize}

The above goals can be formulated mathematically as follows:

\paragraph{(1) Hierarchical Information Flow}

Let $ S_i $ represent the set of agents authorized with security level $ i $, and let $ \ell(v) $ denote the permission level of agent $ v $. For a given piece of information $ I $ assigned security level $ i $:
\begin{equation}
    I \in \mathcal{F}_i \Rightarrow v \in S_i \text{ if and only if } \ell(v) \geq i,
\end{equation}
where $\mathcal{F}_i$ represents the set of information associated with level $i$, $S_i$ represents the set of agents authorized to access information of level $i$, and $\ell(v)$ denotes the permission level of agent $v$.

This condition enforces that information categorized at security level $ i $ can only be accessed by agents whose permission levels are at least $ i $. Consequently, information flow is constrained within the authorized subset of agents, preventing unauthorized access.

\paragraph{(2) Attack Rate Minimization}

Define the probability of a successful attack on agent $ v $ at time step $ t $ as $ P_{\text{attack}}(v, t) $. The objective of AgentSafe is to minimize the overall attack success rate across all agents, expressed as:
\begin{equation}
    \min \sum_{v \in V} \sum_{t=1}^{T} P_{\text{attack}}(v, t),
\end{equation}
where $V$ represents the set of all agents in the system, $T$ represents the time horizon over which the attack rate is evaluated, and $P_{\text{attack}}(v, t)$ denotes the probability of a successful attack on agent $v$ at time $t$.


\subsubsection{Defense Mechanisms}
\
\par
We introduce the defense mechanisms embedded within the AgentSafe framework. The defense mechanisms are categorized into two main components: \textbf{ThreatSieve} and \textbf{HierarCache}. Each component is responsible for mitigating security threats at different levels of the MAS.

\paragraph{ThreatSieve:}

ThreatSieve is a critical mechanism within the \textbf{AgentSafe} framework, designed to prevent unauthorized access and identity impersonation among agents. It ensures secure communication by enforcing strict security defense through authentication and permission validation. Specifically, ThreatSieve operates through two primary functions: \textit{Permission Control} and \textit{Message Legitimacy Evaluation}.

\subparagraph{Permission Control}
ThreatSieve regulates communication between agents based on their permission levels. Communication is permitted only if the sender's permission level is greater than or equal to that of the receiver. This is formalized by the authority verification function \( A(v_i, v_j, t) \), defined as:

\begin{equation}
A(v_i, v_j, t) = 
\begin{cases}
1, & \text{if } \mathcal{L}(I_{i, j}) \leq \ell(v_j) \\
0, & \text{otherwise}
\end{cases},
\end{equation}

where \( \ell(v) \) denotes the permission level of agent \( v \) with other agents, and  $\mathcal{L}$ denotes the security levels of the information $I$ sent by agents.

\subparagraph{Message Legitimacy Evaluation}
ThreatSieve further evaluates the legitimacy of each communication message by verifying the identity of the sender. This process involves extracting identity information from the communication content and validating it through a combination of API calls and a specific program that calculates the similarity between information received and several instructions. The identity extraction process can be mathematically expressed as:

\begin{equation}
\text{ID}_i = E(I, \text{field}_m),
\end{equation}

where \( E \) denotes the extraction of content from \( I \), and \( \text{field}_m \) represents the field containing the identity information. All identities \( \theta \) in \( I \) are extracted by calling an \textbf{API} and an \textbf{LLM} \( l \):

\begin{equation}
\theta = \{\vartheta_1, \dots, \vartheta_n \} = l\left(I, P, C\right),
\end{equation}

where \( P \) is a specific prompt and \( C \) is the context. The identification process is then formalized as:

\begin{equation}
Iv(v_i, v_j) = 
\begin{cases} 
1, & \text{if } \prod_{k=1}^{n} M = 1 \\
0, & \text{if } \prod_{k=1}^{n} M = 0
\end{cases},
\end{equation}

where \( \prod_{k=1}^{n} M(ID_i, \vartheta_k, P^{\prime}, C) \) represents the identification process. \( M(ID_i, \vartheta_k, P^{\prime}, C) = 1 \) indicates that the identity information \( \vartheta_k \) is authentic. If all identities are verified as authentic, \( Iv(v_i, v_j) = 1 \); otherwise, \( Iv(v_i, v_j) = 0 \).

\definecolor{up}{rgb}{0.8, 0, 0.0}
\definecolor{down}{rgb}{0.0, 0.7, 0.0}
\definecolor{right}{rgb}{0.8, 0.7, 0.0}
\begin{table*}[t]
\centering
\caption{Defense Rate comparisons with and without AgentSafe across multiple attack methods and datasets. The number of agents is 7 (6 agents and 1 attacker). The table presents the defense success rates over 10 interaction turns for both the RIOH and WCEI datasets. The results highlight the effectiveness of AgentSafe in mitigating various attack types, such as topology-based attacks including Information Acquisition Based on Topology (IABT) and Authorization Mixup (AM), memory-based attacks including Information Interference (II) and Identity Manipulation (IM) (See Appendix \ref{sec:attacks}).}

\vspace{-1em}
\label{tab:comparison}
\begin{adjustbox}{width=\textwidth}
\begin{tabular}{|llcccccccccc|}
\hline
\rowcolor{cyan!10}
\multicolumn{2}{|l|}{}                                                       & \multicolumn{10}{c|}{Turn}     \\ \hline
\rowcolor{cyan!6}
\multicolumn{2}{|l|}{Attack Method/Dataset} &
  \multicolumn{1}{l}{Turn 5} &
  \multicolumn{1}{l}{Turn 10} &
  \multicolumn{1}{l}{Turn 15} &
  \multicolumn{1}{l}{Turn 20} &
  \multicolumn{1}{l}{Turn 25} &
  \multicolumn{1}{l}{Turn 30} &
  \multicolumn{1}{l}{Turn 35} &
  \multicolumn{1}{l}{Turn 40} &
  \multicolumn{1}{l}{Turn 45} &
  \multicolumn{1}{l|}{Turn 50} \\ \hline
\multicolumn{12}{|l|}{RIOH: \textcolor{gray}{\textit{Describing Privacy Information and Interpersonal Relationships in Common Social Contexts}}}                                                                                          \\ \hline
\multicolumn{1}{|l|}{\multirow{4}{*}{AgentSafe}} & \multicolumn{1}{l|}{\cellcolor{gray!15}IABT} & 
  \cellcolor{gray!15}$80.67$ &
  \cellcolor{gray!15}$73.25_{\textcolor{down}{\downarrow 7.42}}$ &
  \cellcolor{gray!15}$71.76_{\textcolor{down}{\downarrow 1.49}}$ &
  \cellcolor{gray!15}$65.29_{\textcolor{down}{\downarrow 6.47}}$ &
  \cellcolor{gray!15}$52.95_{\textcolor{down}{\downarrow 12.3}}$ &
  \cellcolor{gray!15}$60.42_{\textcolor{up}{\uparrow 7.47}}$ &
  \cellcolor{gray!15}$63.29_{\textcolor{up}{\uparrow 2.87}}$ &
  \cellcolor{gray!15}$58.13_{\textcolor{down}{\downarrow 5.16}}$ &
  \cellcolor{gray!15}$58.47_{\textcolor{up}{\uparrow 0.34}}$ &
  \cellcolor{gray!15}$55.20_{\textcolor{down}{\downarrow 3.27}}$\\
\multicolumn{1}{|l|}{}                           & \multicolumn{1}{l|}{AM}   & 
  \cellcolor{white}$85.93$ &
  \cellcolor{white}$83.25_{\textcolor{down}{\downarrow 2.68}}$ &
  \cellcolor{white}$85.01_{\textcolor{up}{\uparrow 1.76}}$ &
  \cellcolor{white}$83.50_{\textcolor{down}{\downarrow 1.51}}$ &
  \cellcolor{white}$81.25_{\textcolor{down}{\downarrow 2.25}}$ &
  \cellcolor{white}$85.67_{\textcolor{up}{\uparrow 4.42}}$ &
  \cellcolor{white}$86.87_{\textcolor{up}{\uparrow 1.20}}$ &
  \cellcolor{white}$78.13_{\textcolor{down}{\downarrow 8.74}}$ &
  \cellcolor{white}$81.25_{\textcolor{up}{\uparrow 3.12}}$ &
  \cellcolor{white}$82.50_{\textcolor{up}{\uparrow 1.25}}$\\
\multicolumn{1}{|l|}{}                           & \multicolumn{1}{l|}{\cellcolor{gray!15}II}   & 
  \cellcolor{gray!15}$96.88$ &
  \cellcolor{gray!15}$95.83_{\textcolor{down}{\downarrow 1.05}}$ &
  \cellcolor{gray!15}$97.62_{\textcolor{up}{\uparrow 1.79}}$ &
  \cellcolor{gray!15}$91.96_{\textcolor{down}{\downarrow 5.66}}$ &
  \cellcolor{gray!15}$88.33_{\textcolor{down}{\downarrow 3.63}}$ &
  \cellcolor{gray!15}$90.49_{\textcolor{up}{\uparrow 2.16}}$ &
  \cellcolor{gray!15}$85.30_{\textcolor{down}{\downarrow 5.19}}$ &
  \cellcolor{gray!15}$87.68_{\textcolor{up}{\uparrow 2.38}}$ &
  \cellcolor{gray!15}$89.88_{\textcolor{up}{\uparrow 2.20}}$ &
  \cellcolor{gray!15}$88.51_{\textcolor{down}{\downarrow 1.37}}$\\
\multicolumn{1}{|l|}{}                           & \multicolumn{1}{l|}{IM}   & 
  \cellcolor{white}$77.48$ &
  \cellcolor{white}$66.48_{\textcolor{down}{\downarrow 11.0}}$ &
  \cellcolor{white}$65.94_{\textcolor{down}{\downarrow 0.54}}$ &
  \cellcolor{white}$59.56_{\textcolor{down}{\downarrow 6.38}}$ &
  \cellcolor{white}$68.01_{\textcolor{up}{\uparrow 8.45}}$ &
  \cellcolor{white}$55.97_{\textcolor{down}{\downarrow 12.0}}$ &
  \cellcolor{white}$63.92_{\textcolor{up}{\uparrow 7.95}}$ &
  \cellcolor{white}$57.18_{\textcolor{down}{\downarrow 6.74}}$ &
  \cellcolor{white}$53.73_{\textcolor{down}{\downarrow 3.45}}$ &
  \cellcolor{white}$45.83_{\textcolor{down}{\downarrow 7.90}}$\\ \hline
\multicolumn{1}{|l|}{\multirow{4}{*}{w/o AgentSafe}}       & \multicolumn{1}{l|}{\cellcolor{gray!15}IABT} & 
  \cellcolor{gray!15}$34.24$ &
  \cellcolor{gray!15}$22.64_{\textcolor{down}{\downarrow 11.6}}$ &
  \cellcolor{gray!15}$27.02_{\textcolor{up}{\uparrow 4.42}}$ &
  \cellcolor{gray!15}$17.40_{\textcolor{down}{\downarrow 9.62}}$ &
  \cellcolor{gray!15}$26.79_{\textcolor{up}{\uparrow 9.39}}$ &
  \cellcolor{gray!15}$24.56_{\textcolor{down}{\downarrow 2.23}}$ &
  \cellcolor{gray!15}$14.87_{\textcolor{down}{\downarrow 9.69}}$ &
  \cellcolor{gray!15}$27.42_{\textcolor{up}{\uparrow 12.5}}$ &
  \cellcolor{gray!15}$15.07_{\textcolor{down}{\downarrow 13.3}}$ &
  \cellcolor{gray!15}$27.85_{\textcolor{up}{\uparrow 12.7}}$\\
\multicolumn{1}{|l|}{}                           & \multicolumn{1}{l|}{AM}   & 
  \cellcolor{white}$50.32$ &
  \cellcolor{white}$46.88_{\textcolor{down}{\downarrow 3.44}}$ &
  \cellcolor{white}$48.86_{\textcolor{up}{\uparrow 1.98}}$ &
  \cellcolor{white}$46.25_{\textcolor{down}{\downarrow 2.61}}$ &
  \cellcolor{white}$54.31_{\textcolor{up}{\uparrow 8.06}}$ &
  \cellcolor{white}$45.63_{\textcolor{down}{\downarrow 8.68}}$ &
  \cellcolor{white}$49.96_{\textcolor{up}{\uparrow 4.33}}$ &
  \cellcolor{white}$55.63_{\textcolor{up}{\uparrow 5.67}}$ &
  \cellcolor{white}$53.50_{\textcolor{down}{\downarrow 2.13}}$ &
  \cellcolor{white}$55.00_{\textcolor{up}{\uparrow 1.50}}$\\
\multicolumn{1}{|l|}{}                           & \multicolumn{1}{l|}{\cellcolor{gray!15}II}   & 
  \cellcolor{gray!15}$26.88$ &
  \cellcolor{gray!15}$25.63_{\textcolor{down}{\downarrow 1.25}}$ &
  \cellcolor{gray!15}$16.87_{\textcolor{down}{\downarrow 8.76}}$ &
  \cellcolor{gray!15}$19.38_{\textcolor{up}{\uparrow 2.51}}$ &
  \cellcolor{gray!15}$21.25_{\textcolor{up}{\uparrow 1.87}}$ &
  \cellcolor{gray!15}$15.63_{\textcolor{down}{\downarrow 5.62}}$ &
  \cellcolor{gray!15}$19.37_{\textcolor{up}{\uparrow 3.74}}$ &
  \cellcolor{gray!15}$22.50_{\textcolor{up}{\uparrow 3.13}}$ &
  \cellcolor{gray!15}$20.66_{\textcolor{down}{\downarrow 1.84}}$ &
  \cellcolor{gray!15}$14.38_{\textcolor{down}{\downarrow 6.28}}$\\
\multicolumn{1}{|l|}{}                           & \multicolumn{1}{l|}{IM}   & 
  \cellcolor{white}$30.91$ &
  \cellcolor{white}$24.38_{\textcolor{down}{\downarrow 6.53}}$ &
  \cellcolor{white}$26.38_{\textcolor{up}{\uparrow 2.00}}$ &
  \cellcolor{white}$25.63_{\textcolor{down}{\downarrow 0.75}}$ &
  \cellcolor{white}$18.12_{\textcolor{down}{\downarrow 7.51}}$ &
  \cellcolor{white}$24.37_{\textcolor{up}{\uparrow 6.25}}$ &
  \cellcolor{white}$25.83_{\textcolor{up}{\uparrow 1.46}}$ &
  \cellcolor{white}$16.86_{\textcolor{down}{\downarrow 8.97}}$ &
  \cellcolor{white}$22.92_{\textcolor{up}{\uparrow 6.06}}$ &
  \cellcolor{white}$23.13_{\textcolor{up}{\uparrow 0.21}}$\\ \hline
\multicolumn{12}{|l|}{WCEI: \textcolor{gray}{\textit{Describing Privacy Information and Interpersonal Relationships in Corporate Environments}}}                                                                                          \\ \hline
\multicolumn{1}{|l|}{\multirow{4}{*}{AgentSafe}} & \multicolumn{1}{l|}{\cellcolor{gray!15}IABT} & 
  \cellcolor{gray!15}$81.08$ &
  \cellcolor{gray!15}$79.86_{\textcolor{down}{\downarrow 1.22}}$ &
  \cellcolor{gray!15}$74.78_{\textcolor{down}{\downarrow 5.08}}$ &
  \cellcolor{gray!15}$56.90_{\textcolor{down}{\downarrow 17.9}}$ &
  \cellcolor{gray!15}$69.93_{\textcolor{up}{\uparrow 13.0}}$ &
  \cellcolor{gray!15}$58.33_{\textcolor{down}{\downarrow 11.6}}$ &
  \cellcolor{gray!15}$62.36_{\textcolor{up}{\uparrow 4.03}}$ &
  \cellcolor{gray!15}$54.43_{\textcolor{down}{\downarrow 7.93}}$ &
  \cellcolor{gray!15}$53.49_{\textcolor{down}{\downarrow 0.94}}$ &
  \cellcolor{gray!15}$59.51_{\textcolor{up}{\uparrow 6.02}}$\\
\multicolumn{1}{|l|}{}                           & \multicolumn{1}{l|}{AM}   & 
  \cellcolor{white}$88.25$ &
  \cellcolor{white}$84.99_{\textcolor{down}{\downarrow 3.26}}$ &
  \cellcolor{white}$84.53_{\textcolor{down}{\downarrow 0.46}}$ &
  \cellcolor{white}$86.25_{\textcolor{up}{\uparrow 1.72}}$ &
  \cellcolor{white}$82.43_{\textcolor{down}{\downarrow 3.82}}$ &
  \cellcolor{white}$81.25_{\textcolor{down}{\downarrow 1.18}}$ &
  \cellcolor{white}$73.07_{\textcolor{down}{\downarrow 8.18}}$ &
  \cellcolor{white}$86.25_{\textcolor{up}{\uparrow 13.2}}$ &
  \cellcolor{white}$77.11_{\textcolor{down}{\downarrow 9.14}}$ &
  \cellcolor{white}$82.50_{\textcolor{up}{\uparrow 5.39}}$\\
\multicolumn{1}{|l|}{}                           & \multicolumn{1}{l|}{\cellcolor{gray!15}II}   & 
  \cellcolor{gray!15}$87.62$ &
  \cellcolor{gray!15}$81.77_{\textcolor{down}{\downarrow 5.85}}$ &
  \cellcolor{gray!15}$81.88_{\textcolor{up}{\uparrow 0.11}}$ &
  \cellcolor{gray!15}$85.63_{\textcolor{up}{\uparrow 3.75}}$ &
  \cellcolor{gray!15}$76.18_{\textcolor{down}{\downarrow 9.45}}$ &
  \cellcolor{gray!15}$79.17_{\textcolor{up}{\uparrow 2.99}}$ &
  \cellcolor{gray!15}$74.49_{\textcolor{down}{\downarrow 4.68}}$ &
  \cellcolor{gray!15}$73.29_{\textcolor{down}{\downarrow 1.20}}$ &
  \cellcolor{gray!15}$73.05_{\textcolor{down}{\downarrow 0.24}}$ &
  \cellcolor{gray!15}$59.36_{\textcolor{down}{\downarrow 13.7}}$\\
\multicolumn{1}{|l|}{}                           & \multicolumn{1}{l|}{IM}   & 
  \cellcolor{white}$71.72$ &
  \cellcolor{white}$57.92_{\textcolor{down}{\downarrow 13.8}}$ &
  \cellcolor{white}$64.95_{\textcolor{up}{\uparrow 7.03}}$ &
  \cellcolor{white}$68.13_{\textcolor{up}{\uparrow 3.18}}$ &
  \cellcolor{white}$66.03_{\textcolor{down}{\downarrow 2.1}}$ &
  \cellcolor{white}$56.25_{\textcolor{down}{\downarrow 9.78}}$ &
  \cellcolor{white}$67.63_{\textcolor{up}{\uparrow 11.4}}$ &
  \cellcolor{white}$65.63_{\textcolor{down}{\downarrow 2.00}}$ &
  \cellcolor{white}$59.95_{\textcolor{down}{\downarrow 5.68}}$ &
  \cellcolor{white}$43.75_{\textcolor{down}{\downarrow 16.2}}$\\ \hline
\multicolumn{1}{|l|}{\multirow{4}{*}{w/o AgentSafe}}       & \multicolumn{1}{l|}{\cellcolor{gray!15}IABT} & 
  \cellcolor{gray!15}$25.51$ &
  \cellcolor{gray!15}$20.21_{\textcolor{down}{\downarrow 5.30}}$ &
  \cellcolor{gray!15}$28.8_{\textcolor{up}{\uparrow 8.59}}$ &
  \cellcolor{gray!15}$20.45_{\textcolor{down}{\downarrow 8.35}}$ &
  \cellcolor{gray!15}$30.95_{\textcolor{up}{\uparrow 10.5}}$ &
  \cellcolor{gray!15}$13.60_{\textcolor{down}{\downarrow 17.6}}$ &
  \cellcolor{gray!15}$22.87_{\textcolor{up}{\uparrow 9.27}}$ &
  \cellcolor{gray!15}$27.09_{\textcolor{up}{\uparrow 4.22}}$ &
  \cellcolor{gray!15}$28.62_{\textcolor{up}{\uparrow 1.53}}$ &
  \cellcolor{gray!15}$21.66_{\textcolor{down}{\downarrow 6.96}}$\\
\multicolumn{1}{|l|}{}                           & \multicolumn{1}{l|}{AM}   & 
  \cellcolor{white}$52.52$ &
  \cellcolor{white}$46.25_{\textcolor{down}{\downarrow 6.27}}$ &
  \cellcolor{white}$53.26_{\textcolor{up}{\uparrow 7.01}}$ &
  \cellcolor{white}$45.63_{\textcolor{down}{\downarrow 7.63}}$ &
  \cellcolor{white}$53.87_{\textcolor{up}{\uparrow 8.24}}$ &
  \cellcolor{white}$52.50_{\textcolor{down}{\downarrow 1.37}}$ &
  \cellcolor{white}$55.33_{\textcolor{up}{\uparrow 2.83}}$ &
  \cellcolor{white}$44.38_{\textcolor{down}{\downarrow 10.9}}$ &
  \cellcolor{white}$48.68_{\textcolor{up}{\uparrow 4.3}}$ &
  \cellcolor{white}$45.99_{\textcolor{down}{\downarrow 2.69}}$\\
\multicolumn{1}{|l|}{}                           & \multicolumn{1}{l|}{\cellcolor{gray!15}II}   & 
  \cellcolor{gray!15}$29.56$ &
  \cellcolor{gray!15}$24.38_{\textcolor{down}{\downarrow 5.18}}$ &
  \cellcolor{gray!15}$24.82_{\textcolor{up}{\uparrow 0.44}}$ &
  \cellcolor{gray!15}$19.38_{\textcolor{down}{\downarrow 5.44}}$ &
  \cellcolor{gray!15}$30.84_{\textcolor{up}{\uparrow 11.5}}$ &
  \cellcolor{gray!15}$23.12_{\textcolor{down}{\downarrow 7.72}}$ &
  \cellcolor{gray!15}$11.92_{\textcolor{down}{\downarrow 11.2}}$ &
  \cellcolor{gray!15}$12.50_{\textcolor{up}{\uparrow 0.58}}$ &
  \cellcolor{gray!15}$18.19_{\textcolor{up}{\uparrow 5.69}}$ &
  \cellcolor{gray!15}$18.13_{\textcolor{down}{\downarrow 0.06}}$\\
\multicolumn{1}{|l|}{}                           & \multicolumn{1}{l|}{IM}   & 
  \cellcolor{white}$23.52$ &
  \cellcolor{white}$21.25_{\textcolor{down}{\downarrow 2.27}}$ &
  \cellcolor{white}$31.87_{\textcolor{up}{\uparrow 10.6}}$ &
  \cellcolor{white}$22.50_{\textcolor{down}{\downarrow 9.37}}$ &
  \cellcolor{white}$37.53_{\textcolor{up}{\uparrow 15.0}}$ &
  \cellcolor{white}$23.13_{\textcolor{down}{\downarrow 14.4}}$ &
  \cellcolor{white}$20.80_{\textcolor{down}{\downarrow 2.33}}$ &
  \cellcolor{white}$21.25_{\textcolor{up}{\uparrow 0.45}}$ &
  \cellcolor{white}$19.02_{\textcolor{down}{\downarrow 2.23}}$ &
  \cellcolor{white}$21.25_{\textcolor{up}{\uparrow 2.23}}$\\ \hline
\end{tabular}
\end{adjustbox}
\end{table*}

\paragraph{HierarCache:}

HierarCache is a critical component of the \textbf{AgentSafe} framework, designed to manage hierarchical information storage and ensure information security within agent memory. It organizes memory into multiple layers, each corresponding to a specific security level, and includes an additional "junk" memory layer for storing irrelevant or harmful information. This structure ensures that sensitive information is not leaked while maintaining the integrity of stored data.

\subparagraph{Hierarchical Storage Mechanism}
The memory update function \( U(v_i, v_j, m, \ell) \) is defined for a message \( m \) sent from agent \( v_i \) to agent \( v_j \), where \( \ell \) is the security level of the message:

\begin{equation}
U(v_i, v_j, m, \ell) =
\begin{cases}
f_{\ell}(m), & \text{if } \text{Vd} = 1 \\
f_{\text{junk}}(m), & \text{otherwise}
\end{cases}
\end{equation}

where \( f_{\ell}(m) \) denotes the operation of adding message \( m \) to the memory set \( M_{\ell} \), and \( f_{\text{junk}}(m) \) denotes the operation of adding \( m \) to the junk memory set \( M_{\text{junk}} \). The validity condition \( \text{Vd}(v_i, m, \ell) = 1 \) is defined as \( \ell(v_i) \geq \ell \land D(m) = 1 \), where \( \ell(v_i) \) represents the permission level of the sending agent \( v_i \), and \( D(m) \) is the detection function that assesses the validity of message \( m \).

\subparagraph{Detection Function \( D(m) \)}
The validity of a message \( m \) is determined by comparing it against a set of verification criteria defined in an instruction library \( \mathcal{C} \). Each criterion \( m_i \) is a natural language description, and the similarity between \( m \) and \( m_i \) is calculated using a vector semantic similarity function \( \text{Sim}(m, m_i) \). The detection function \( D(m) \) is defined as:

\begin{equation}
D(m) = 
\begin{cases} 
1, & \text{if } \sum_{i=1}^{n} \delta(m, m_i) = n \\
0, & \text{if } \sum_{i=1}^{n} \delta(m, m_i) < n
\end{cases},
\end{equation}

where \( \delta(m, m_i) \) is an indicator function that can be expressed as:

\begin{equation}
\delta(m, m_i) = \mathbb{I}(\text{Sim}(m, m_i) > \theta).
\end{equation}

\( \text{Sim}(m, m_i) \) represents the similarity between \( m \) and \( m_i \), such as cosine similarity or Euclidean distance, and \( \theta \) is a predefined similarity threshold. If the similarity exceeds \( \theta \), the message \( m \) is considered to satisfy the \( i \)-th verification criterion.

\subparagraph{Periodic Detection and Isolation Mechanism}
To ensure the correctness of stored information, HierarCache employs a periodic detection mechanism that inspects and isolates false information. The detection process is formalized as:

\begin{equation}
R(v_j, t) = l(\rho^t),
\end{equation}

where \( l \) is the language model used for reflection, and \( \rho^t \) is a prompt designed to encourage the model to reflect on the information. The prompt \( \rho^t \) is defined as:

\begin{equation}
\rho^t = \{ \text{reflection}, \mathcal{C}, M_{\text{junk}}^t \},
\end{equation}

where \( \mathcal{C} \) is the instruction library, and \( M_{\text{junk}}^t \) represents the junk memory at time \( t \). The set of false information \( F_{\ell} \) identified during the detection process is expressed as:

\begin{equation}
F_{\ell} = \{ m \mid R(v_j, t) = \text{"junk"} \}.
\end{equation}

After detection, if \( F_{\ell} \) is a subset of \( M_{\ell} \), then \( M_{\ell} \) is updated by removing \( F_{\ell} \) from it (\( M_{\ell} \leftarrow M_{\ell} \setminus F_{\ell} \)). Otherwise, all sets \( F_{\ell} \) in the collection \( \mathcal{F} \) are added to \( M_{\text{junk}} \) (\( M_{\text{junk}} \leftarrow M_{\text{junk}} \cup \bigcup_{F_{\ell} \in \mathcal{F}} F_{\ell} \)).


This mechanism ensures that false information is removed from secure memory levels and transferred to junk memory, thereby maintaining the integrity of the hierarchical storage system.

%% file: 5_experiment.tex
\SetAlgoNoEnd

\section{Experiment}
To thoroughly investigate the defense mechanisms of \textbf{AgentSafe} under diverse attack vectors and its performance across various real-world applications, we structured our experiments around several key research questions. These experiments aim to evaluate, answer, and summarize the resilience and effectiveness of AgentSafe in practical deployment scenarios. 

\begin{itemize}[leftmargin=*]
    \item \textbf{RQ1:} How effective is AgentSafe in multi-agent systems under multi-round interactions?
    \item \textbf{RQ2:} Does AgentSafe defend against multi-round attacks across different LLMs?
    \item \textbf{RQ3:} How does system complexity impact the performance of AgentSafe in maintaining data integrity?
    \item \textbf{RQ4:} How does AgentSafe perform in defending against attacks in MASs with varying topological structures?
\end{itemize}

\subsection{Experimental Setups}
\textbf{Datasets.} Previous datasets lacked both interpersonal relationships and multi-level privacy information. To address this gap and simulate diverse human relationships and privacy levels, we introduce the Relationship and Information of Human (RIOH) dataset and the Whole Company Employee Information (WCEI) dataset. The structure and details of these two datasets can be found in Appendix \ref{appendix:datasets} for further reference.

\textbf{Models and Metrics.} To comprehensively evaluate the performance of AgentSafe in hierarchical information handling and defense across various large language models, we utilize APIs including \llmname{Llama 3.2}\footnote{\url{https://llama.meta.com/}}, \llmname{GPT-3.5-Turbo}\footnote{\url{https://openai.com/research/gpt-3-5-turbo}}, \llmname{GPT-4o}\footnote{\url{https://openai.com/research/gpt-4o}}, \llmname{GPT-4o-mini}\footnote{\url{https://openai.com/research/gpt-4o-mini}}, and \llmname{GPT-4}\footnote{\url{https://openai.com/research/gpt-4}}.
In our metrics, the Defense Rate is defined as the ratio of the total number of successfully defended attacks by all agents to the total number of attacks. CSI represents the cosine similarity between the output and the ground-truth information. Additionally, the Cosine Similarity Rate (CSR) is the ratio of the cosine similarity between outputs without AgentSafe and with AgentSafe, relative to the original message.

\begin{figure}[t]
  \centering
  \includegraphics[width=1\linewidth]{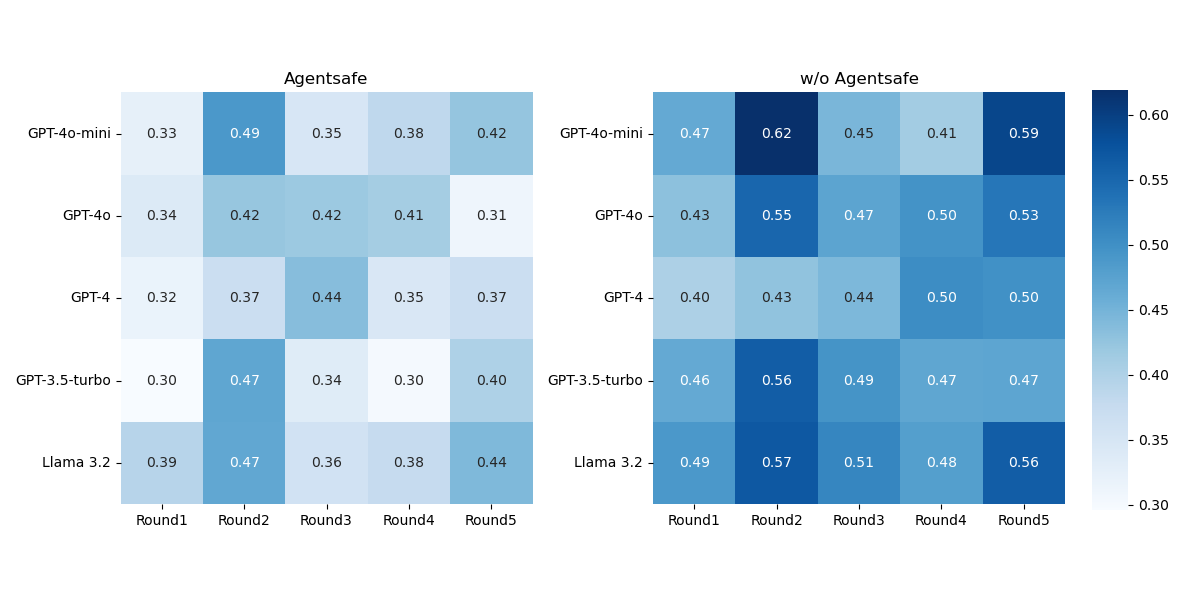}
  \vspace{-1.4em}
  \caption{his figure presents the results of multiple-round TBA attacks on the AgentSafe and non-AgentSafe frameworks in different API environments. The higher the value, the closer the output result is to the real data, which means the better the attack effect. On the contrary, the defense effect is better. }

  \label{fig:heatmap}
\end{figure}
 
 \subsection{Defense and Ablation Results (RQ1)}
To validate our framework and address RQ1, we evaluate the performance of AgentSafe across different datasets under four distinct attack methods as the number of communication rounds increases (Table ~\ref{tab:comparison}). Our findings reveal three key insights:

\textbf{Obs\textcircled{1}. Enhanced Defense Capabilities of AgentSafe:}
As shown in Table 1, AgentSafe consistently demonstrates superior defense performance across all four attack types and different datasets. In contrast, the baseline LLM (w/o AgentSafe) exhibits significantly lower defense success rates. For instance, under the topology-based attack (TBA), AgentSafe achieves an 80.67\% success rate at turn 5, whereas the baseline LLM only reaches 34.24\%. \textit{This highlights the weaker defense of the case without AgentSafe, further validating that the hierarchical defense mechanism of AgentSafe enables a more robust security posture.}

\textbf{Obs\textcircled{2}. Sustained Performance Over Multiple Rounds of Interaction:}
We evaluate AgentSafe across 5 to 50 rounds of interaction under each attack type. While both AgentSafe and the baseline LLM show some decline in defense success rates over time, AgentSafe maintains significantly higher performance levels, even as the number of turns increases. Notably, in the topology-based attack scenario, AgentSafe’s defense success rate drops to 55.20\% at turn 50, which is still higher than the baseline LLM's peak performance. This demonstrates the enduring effectiveness of AgentSafe in sustaining high defense success rates across multiple rounds of adversarial interactions.


\begin{figure}[t]
  \centering
  \includegraphics[width=1\linewidth]{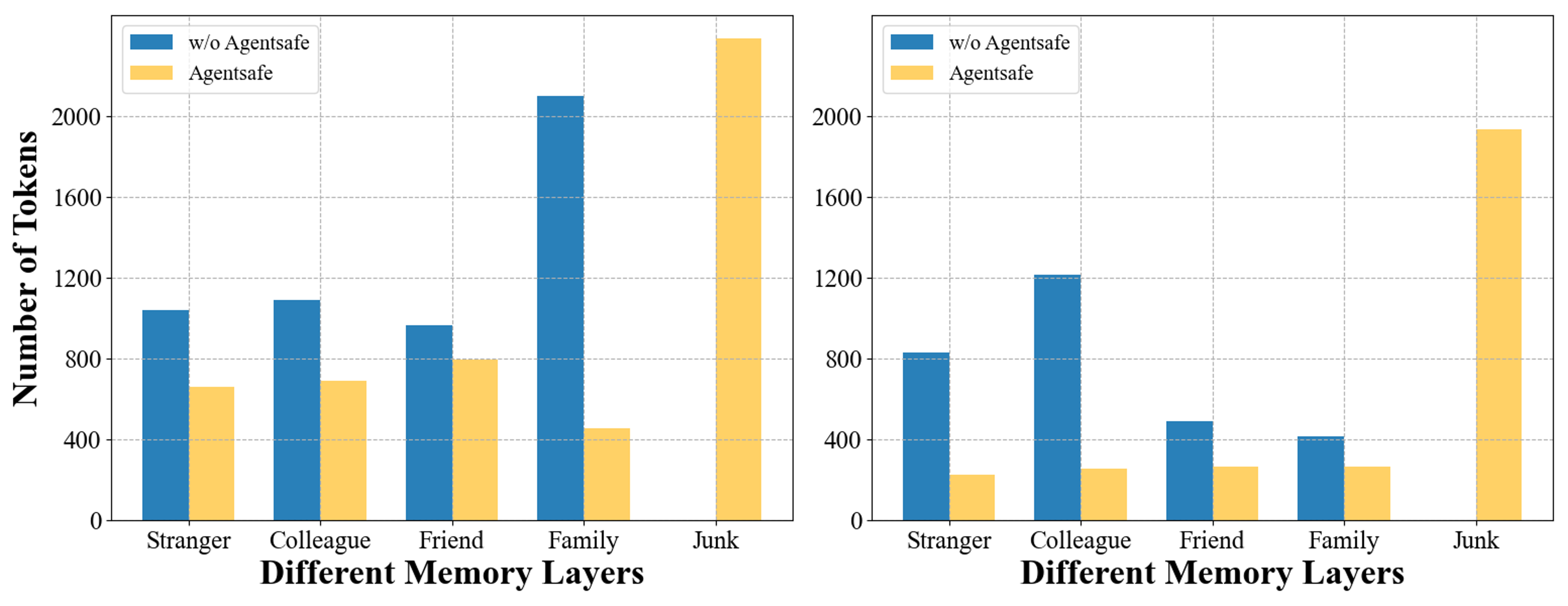}
  \vspace{-1.4em}
  \caption{The left and right figures show the comparison of token consumption with and without AgentSafe under topology-based and memory-based attacks respectively.}

  \label{fig:token}
\end{figure}

\textbf{Obs\textcircled{3}. Efficiency Improvement with Overhead Reduction:}
In scenarios with numerous communication turns, irrelevant data can impede agent interactions. By filtering out useless or harmful information, AgentSafe significantly reduces system overhead and ensures smooth interactions during communication, enhancing overall interaction efficiency. As shown in Figure~\ref{fig:token}, under different attacks with many communication turns, AgentSafe significantly reduces token consumption. For topology-based attacks, there is a 60\% reduction in tokens, and for memory-based attacks, a 75\% reduction occurs. This not only cuts storage load but also boosts retrieval efficiency for the whole system. Appendix \ref{sec: overhead} documents the detailed analysis and derivation regarding how AgentSafe reduces the overall system overhead.

\subsection{Multi-round Attacks Analysis (RQ2)}

\begin{figure}[t]
  \centering
  \includegraphics[width=1\linewidth]{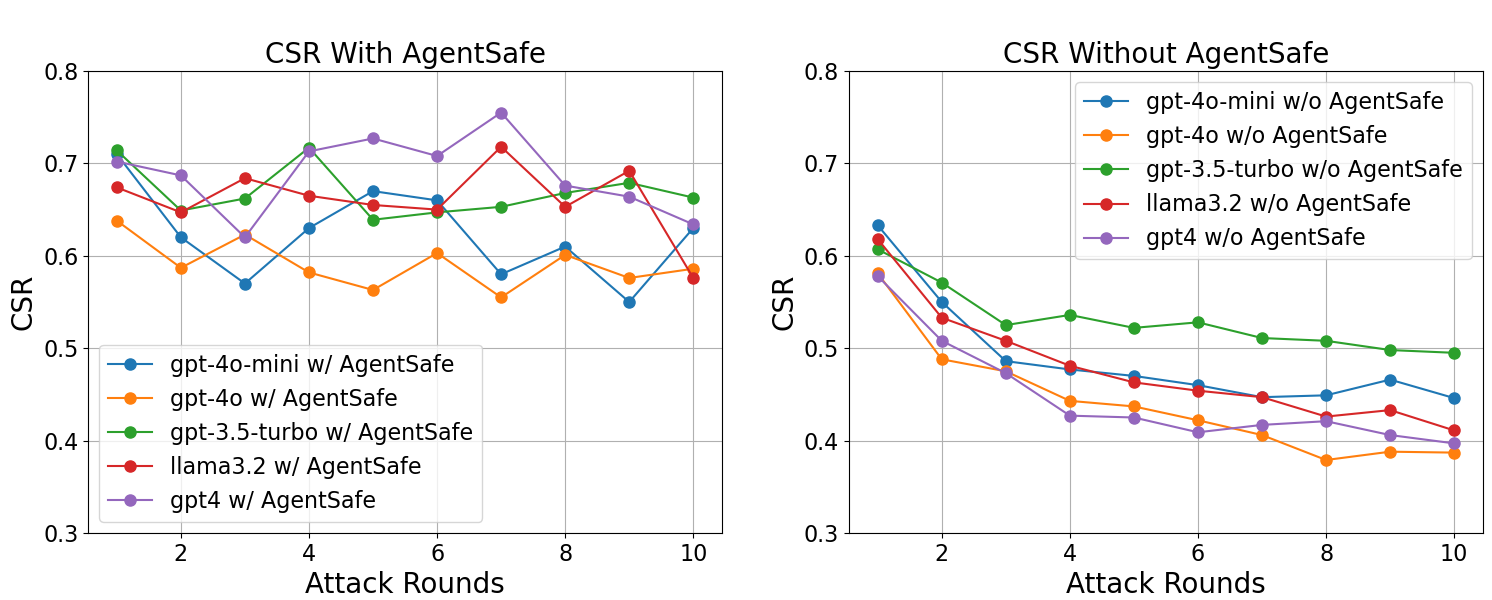}
  \vspace{-1.4em}
  \caption{This figure shows the results of using multiple rounds of MBA to attack the AgentSafe framework and the non-AgentSafe framework across different LLMs.}  

  \label{fig:RQ2}
\end{figure}

To answer \textbf{RQ2}, we conduct multi-round attack experiments comparing the performance of the AgentSafe framework and the baseline (without AgentSafe) across different LLM environments. The attack methods include both multi-round topology-based attacks (TBA) and memory-based attacks (MBA). We use cosine similarity to measure the closeness between the output data or memory data and the original data. Figure \ref{fig:heatmap} and Figure \ref{fig:RQ2} present the experimental results under different APIs and attack models.

\textbf{Obs\textcircled{1}. Significant Reduction in the Impact of Multi-round Attacks:} 
As shown in Figure \ref{fig:heatmap}, AgentSafe significantly mitigates the effect of topology-based attacks (TBA), maintaining lower cosine similarity compared to the baseline. For example, cosine similarity after the fifth round is 0.44 with AgentSafe in LLaMA 3.2, compared to 0.56 without it, demonstrating its protective effect. Similarly, in memory-based attacks (MBA) (Figure \ref{fig:RQ2}), AgentSafe maintains higher cosine similarity (0.65-0.75) after 10 rounds, much higher than the 0.4 observed without it, showing its effectiveness in protecting information integrity under sustained attacks.

\textbf{Obs\textcircled{2}. Consistent and Robust Performance across Different LLMs:} 
As Figure \ref{fig:heatmap} and Figure \ref{fig:RQ2} show, AgentSafe exhibits robust defense capabilities in different environments. Its cosine similarity remains high across all settings. For example, in the \llmname{GPT-4o-mini} environment, after multiple attack rounds, the cosine similarity is 0.42 with AgentSafe, versus 0.59 without it. In the MBA attack experiments with \llmname{LLaMA 3.2} (Figure \ref{fig:RQ2}), the defense efficiency is especially notable. After the 10th round, the cosine similarity with AgentSafe hovers around 0.7, while the baseline without it drops below 0.4. These results indicate AgentSafe can consistently maintain high-level defense performance across various LLMs, safeguarding data from severe attack impacts.

\definecolor{up}{rgb}{0.8, 0, 0.0}
\definecolor{down}{rgb}{0.0, 0.0, 0.7}
\definecolor{right}{rgb}{0.8, 0.7, 0.0}
\begin{table*}[t]
\centering
\caption{CSI comparisons in multi-agent systems with different topological structures and number of agents, comparing with and without AgentSafe.}

\vspace{-1em}
\label{tab:new_table}
\begin{adjustbox}{width=\textwidth}
\begin{tabular}{|llccccccc|}
\hline
\rowcolor{orange!30}
\multicolumn{2}{|l|}{}                                                       & \multicolumn{7}{c|}{Number of Agents}     \\ \hline
\rowcolor{orange!10}
\multicolumn{2}{|l|}{Topological Strutures} &
  \multicolumn{1}{l}{4} &
  \multicolumn{1}{l}{5} &
  \multicolumn{1}{l}{6} &
  \multicolumn{1}{l}{7} &
  \multicolumn{1}{l}{8} &
  \multicolumn{1}{l}{9} &
  \multicolumn{1}{l|}{10} \\ \hline

\multicolumn{1}{|l|}{\multirow{2}{*}{Chain}}                           & \multicolumn{1}{l|}{\cellcolor{gray!15}AgentSafe}   & 
  \cellcolor{gray!15}$44.34_{\textcolor{down}{\downarrow 10.12}}$ &
  \cellcolor{gray!15}$34.62_{\textcolor{down}{\downarrow 37.74}}$ &
  \cellcolor{gray!15}$47.11_{\textcolor{down}{\downarrow 11.18}}$ &
  \cellcolor{gray!15}$42.09_{\textcolor{down}{\downarrow 19.50}}$ &
  \cellcolor{gray!15}$40.12_{\textcolor{down}{\downarrow 10.07}}$ &
  \cellcolor{gray!15}$39.89_{\textcolor{down}{\downarrow 24.32}}$ &
  \cellcolor{gray!15}$42.72_{\textcolor{down}{\downarrow 27.70}}$\\
\multicolumn{1}{|l|}{}                           & \multicolumn{1}{l|}{w/o AgentSafe}   & 
  \cellcolor{white}$54.46$ &
  \cellcolor{white}$72.36$ &
  \cellcolor{white}$58.30$ &
  \cellcolor{white}$61.60$ &
  \cellcolor{white}$50.18$ &
  \cellcolor{white}$64.22$ &
  \cellcolor{white}$70.42$\\ \hline

\multicolumn{1}{|l|}{\multirow{2}{*}{Cycle}}       & \multicolumn{1}{l|}{\cellcolor{gray!15}AgentSafe}   & 
  \cellcolor{gray!15}$47.82_{\textcolor{down}{\downarrow 2.36}}$ &
  \cellcolor{gray!15}$34.69_{\textcolor{down}{\downarrow 9.42}}$ &
  \cellcolor{gray!15}$46.97_{\textcolor{down}{\downarrow 8.27}}$ &
  \cellcolor{gray!15}$37.35_{\textcolor{down}{\downarrow 23.86}}$ &
  \cellcolor{gray!15}$44.72_{\textcolor{down}{\downarrow 5.56}}$ &
  \cellcolor{gray!15}$45.82_{\textcolor{down}{\downarrow 31.85}}$ &
  \cellcolor{gray!15}$46.51_{\textcolor{down}{\downarrow 19.44}}$\\
\multicolumn{1}{|l|}{}                           & \multicolumn{1}{l|}{w/o AgentSafe}   & 
  \cellcolor{white}$50.18$ &
  \cellcolor{white}$44.11$ &
  \cellcolor{white}$55.24$ &
  \cellcolor{white}$61.21$ &
  \cellcolor{white}$50.28$ &
  \cellcolor{white}$77.67$ &
  \cellcolor{white}$65.95$\\ \hline

\multicolumn{1}{|l|}{\multirow{2}{*}{Binary Tree}}                           & \multicolumn{1}{l|}{\cellcolor{gray!15}AgentSafe}   & 
  \cellcolor{gray!15}$43.63_{\textcolor{down}{\downarrow 15.88}}$ &
  \cellcolor{gray!15}$42.06_{\textcolor{down}{\downarrow 20.21}}$ &
  \cellcolor{gray!15}$45.53_{\textcolor{down}{\downarrow 22.06}}$ &
  \cellcolor{gray!15}$47.66_{\textcolor{down}{\downarrow 19.55}}$ &
  \cellcolor{gray!15}$43.98_{\textcolor{down}{\downarrow 17.00}}$ &
  \cellcolor{gray!15}$43.13_{\textcolor{down}{\downarrow 31.66}}$ &
  \cellcolor{gray!15}$41.02_{\textcolor{down}{\downarrow 17.21}}$\\
\multicolumn{1}{|l|}{}                           & \multicolumn{1}{l|}{w/o AgentSafe}   & 
  \cellcolor{white}$59.51$ &
  \cellcolor{white}$62.27$ &
  \cellcolor{white}$67.61$ &
  \cellcolor{white}$63.51$ &
  \cellcolor{white}$60.98$ &
  \cellcolor{white}$74.79$ &
  \cellcolor{white}$58.23$\\ \hline

\multicolumn{1}{|l|}{\multirow{2}{*}{Complete Graph}}       & \multicolumn{1}{l|}{\cellcolor{gray!15}AgentSafe}   & 
  \cellcolor{gray!15}$44.31_{\textcolor{down}{\downarrow 13.41}}$ &
  \cellcolor{gray!15}$32.66_{\textcolor{down}{\downarrow 20.83}}$ &
  \cellcolor{gray!15}$41.98_{\textcolor{down}{\downarrow 11.06}}$ &
  \cellcolor{gray!15}$44.55_{\textcolor{down}{\downarrow 3.30}}$ &
  \cellcolor{gray!15}$44.52_{\textcolor{down}{\downarrow 15.38}}$ &
  \cellcolor{gray!15}$44.86_{\textcolor{down}{\downarrow 16.97}}$ &
  \cellcolor{gray!15}$40.60_{\textcolor{down}{\downarrow 25.20}}$\\
\multicolumn{1}{|l|}{}                           & \multicolumn{1}{l|}{w/o AgentSafe}   & 
  \cellcolor{white}$57.72$ &
  \cellcolor{white}$53.49$ &
  \cellcolor{white}$53.04$ &
  \cellcolor{white}$47.85$ &
  \cellcolor{white}$59.90$ &
  \cellcolor{white}$61.83$ &
  \cellcolor{white}$65.80$\\ \hline

\end{tabular}
\end{adjustbox}
\end{table*}

\subsection{System Complexity on AgentSafe (RQ3)}

\begin{figure}[t]
  \centering
  \includegraphics[width=1\linewidth]{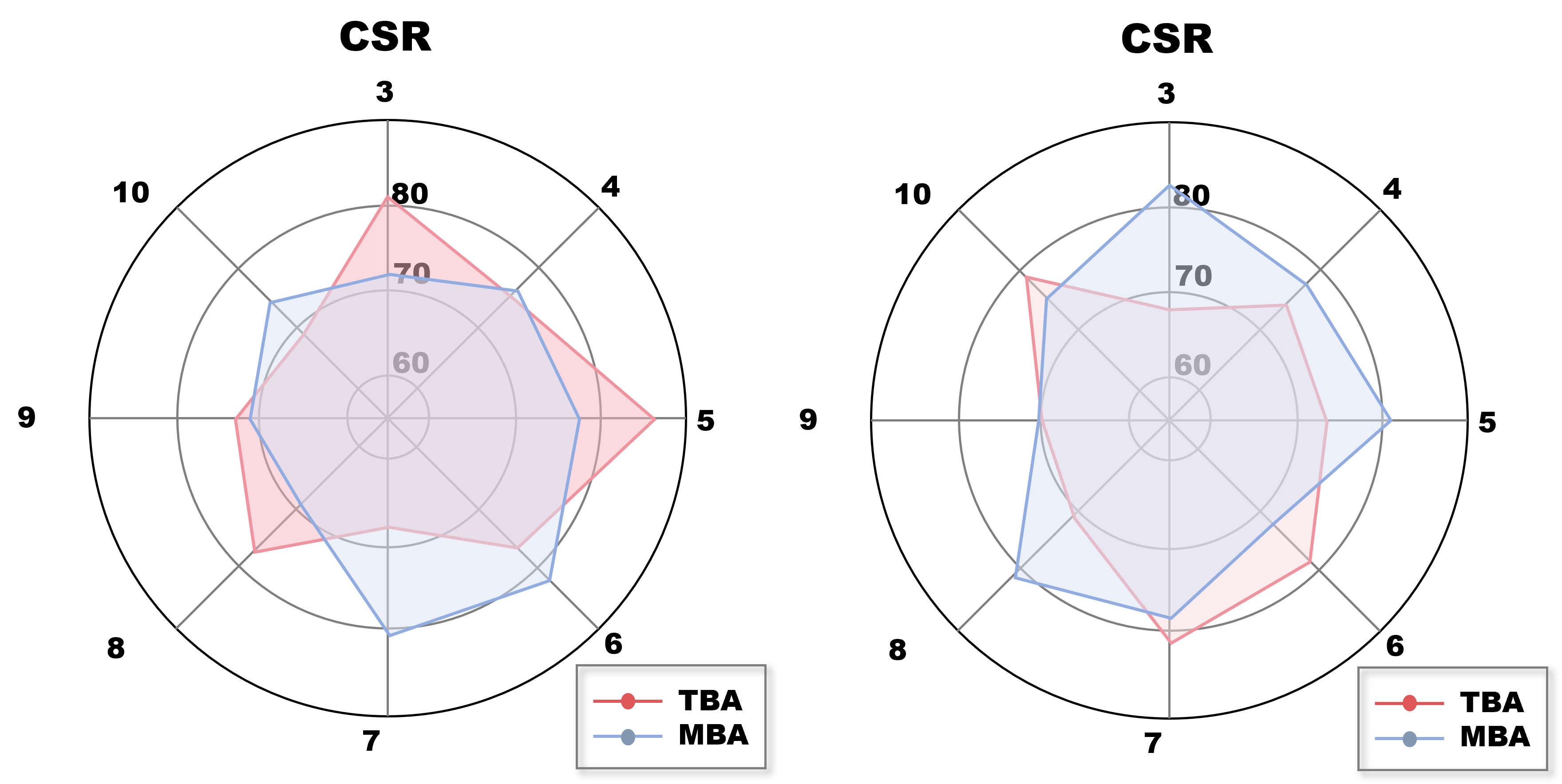}
  \vspace{-1.4em}
  \caption{\textbf{\textit{Left:}} The impact of the number of agents on CSR. \textbf{\textit{Right:}} The relationship between information complexity and CSR.}
  \vspace{-1.4em}
  \label{fig:RQ3r}
\end{figure}

To investigate \textbf{RQ3}, we conduct experiments focusing on two aspects of system complexity: the complexity of memory information and the number of agents in the system. We gradually increase both the memory information complexity and the number of agents to evaluate their effects on the performance. The results are shown in Figure~\ref{fig:RQ3r}.

\textbf{Obs\textcircled{1}. Scalability of AgentSafe in MAS:} 
As shown in Figure~\ref{fig:RQ3r} (left), AgentSafe also scales effectively as the number of agents in the system increases. The CSR remains stable, ranging between 0.68 and 0.85, regardless of whether the attack is topology-based or memory-based. This highlights the ability of AgentSafe to maintain high performance as the system complexity increases in terms of the number of agents. Importantly, there is no significant decrease in the system’s performance, even as the number of agents grows, further validating the scalability of AgentSafe for large-scale, distributed MAS deployments.

\textbf{Obs\textcircled{2}. Limited Impact of Information Complexity on the Performance of AgentSafe:} 
The results, depicted in Figure~\ref{fig:RQ3r} (right), show that the complexity of memory information has a minimal impact on the performance of AgentSafe. The Cosine Similarity Rate (CSR) remains consistent, around 0.67 to 0.82, across different levels of information complexity for both topology-based and memory-based attacks. This demonstrates that AgentSafe is resilient to variations in data complexity and continues to maintain high integrity in the output. This robustness indicates that AgentSafe can handle diverse real-world scenarios where the complexity of the input may vary without significant performance degradation.

\subsection{Topological Structures Analysis (RQ4)}

To answer \textbf{RQ4}, we conduct experiments to evaluate how AgentSafe performs in defending against attacks in MAS with different topological structures. We focus on common topological structures within contemporary MAS, namely the chain, cycle, binary tree, and complete graph, which are widely adopted in MAS research\cite{yu2024netsafe}. In these experiments, we vary the number of agents from 4 to 10. The results are presented in Table \ref{tab:new_table}.

\textbf{Obs\textcircled{1}. Consistent Superior Defense across Topologies:}
As shown in Table \ref{tab:new_table}, regardless of the topological structure of MAS, AgentSafe consistently outperforms the non-AgentSafe case. For the Cycle topology, when \(num\_agents = 9\), the CSI with AgentSafe is  45.82. However, CSI reaches 77.67 without AgentSafe. This consistent pattern across different topologies highlights the effectiveness of AgentSafe's defense mechanism in diverse MAS architectures. It validates that AgentSafe can be effectively applied in real-world MAS with various topological configurations, providing reliable protection against attacks.

\textbf{Obs\textcircled{2}. Resilient Performance with Increasing Agent Numbers:}
In MAS across various topologies, AgentSafe keeps the CSI stable between 35 - 50 regardless of agent count. For example, in the Chain topology, as \(num\_agents\) rises from 4 to 10, it stays within this range. However, without AgentSafe, the CSI increases notably with more agents. In the Chain topology, at 4 agents, the CSI is 54.46, and when \(num\_agents\) reaches 10, it jumps to 70.42. The gap between the two cases widens as agent numbers grow. This shows AgentSafe performs consistently and is more effective in protecting against attacks in complex, high-overhead scenarios.